\newif\ifisTR
\isTRfalse

\documentclass{article}

\usepackage{microtype}
\usepackage{graphicx}
\usepackage{subcaption}
\usepackage{booktabs} 
\usepackage{multirow}
\usepackage{array} 
\newcolumntype{C}{>{\centering\arraybackslash}p{1.5em}}

\usepackage{hyperref}

\usepackage[accepted]{./icml-2026/icml2026}

\usepackage{amsmath}
\usepackage{amssymb}
\usepackage{mathtools}
\usepackage{amsthm}

\usepackage[capitalize,noabbrev]{cleveref}

\usepackage[textsize=tiny]{todonotes}

\usepackage{amsmath,amsfonts,bm}
\usepackage{hyperref}
\usepackage{array} 

\usepackage{tikz}
\usepackage{pict2e}
\usepackage{subcaption}
\usetikzlibrary{arrows.meta,positioning,fit,calc,backgrounds}
\usepackage{xcolor}

\hypersetup{
    urlcolor = magenta
}

\def\eqref#1{equation~\ref{#1}}
\def\Eqref#1{Equation~\ref{#1}}

\def\1{\bm{1}}

\def\rmH{{\mathbf{H}}}

\def\vtheta{{\bm{\theta}}}

\DeclareMathAlphabet{\mathsfit}{\encodingdefault}{\sfdefault}{m}{sl}
\SetMathAlphabet{\mathsfit}{bold}{\encodingdefault}{\sfdefault}{bx}{n}

\newcommand{\R}{\mathbb{R}}

\newcommand{\Dc}{\mathcal D}

\newcommand{\Bc}{\mathcal B}

\newcommand{\nres}{n_\textup{res}}
\newcommand{\nbc}{n_{\textup{bc}}}

\usepackage[utf8]{inputenc}
\usepackage{graphicx}

\usepackage{enumitem}
\usepackage{amsmath}
\usepackage{amsfonts}
\usepackage{dcolumn} 
\usepackage{tabu}
\usepackage{array}
\usepackage{xspace}
\usepackage{makecell}
\usepackage{amsthm}
\usepackage{algorithmic}

\usepackage{colortbl}
\usepackage{booktabs, multirow} 
\usepackage{soul}
\usepackage{changepage,threeparttable} 

\usepackage[dvipsnames]{xcolor}
\usepackage{color, colortbl}
\usepackage{multirow}
\usepackage{enumitem}
\usepackage{placeins}
\usepackage{subcaption}
\usepackage{microtype}
\usepackage{graphicx}
\usepackage{booktabs} 
\usepackage{multirow}
\usepackage{enumitem}
\usepackage{graphicx}
\usepackage{color}
\usepackage{xcolor}
\usepackage{colortbl}
\definecolor{dullGray}{HTML}{F3F3F3}
\definecolor{smallColorT}{HTML}{fff1e6}
\definecolor{largeColorT}{HTML}{edf2fb}

\usepackage{amsmath}
\usepackage{amssymb}
\usepackage{mathtools}
\usepackage{amsthm}

\usepackage[capitalize,noabbrev]{cleveref}

\theoremstyle{plain}

\theoremstyle{definition}

\theoremstyle{remark}

\usepackage[textsize=tiny]{todonotes}

\icmltitlerunning{Unveiling Multi-regime Patterns in SciML: Distinct Failure Modes and Regime-specific Optimization}

\begin{document}

\twocolumn[
\icmltitle{Unveiling Multi-regime Patterns in SciML: Distinct Failure Modes \\and Regime-specific Optimization}



\icmlsetsymbol{equal}{*}
\icmlsetsymbol{corr}{\ddag}

\begin{icmlauthorlist}
\icmlauthor{Yuxin Wang}{equal,dartmouth}
\icmlauthor{Yuanzhe Hu}{equal,ucsd}
\icmlauthor{Xiaokun Zhong}{equal,ucb}
\icmlauthor{Xiaopeng Wang}{equal,dartmouth}
\icmlauthor{Haiquan Lu}{nus}
\icmlauthor{Tianyu Pang}{dartmouth}
\icmlauthor{Michael W. Mahoney}{ucb,lbl,icsi}
\icmlauthor{Yujun Yan}{dartmouth}
\icmlauthor{Pu Ren}{corr,lbl}
\icmlauthor{Yaoqing Yang}{corr,dartmouth}

\end{icmlauthorlist}

\icmlaffiliation{dartmouth}{Dartmouth College}
\icmlaffiliation{ucsd}{University of California, San Diego}
\icmlaffiliation{ucb}{University of California, Berkeley}
\icmlaffiliation{nus}{National University of Singapore}
\icmlaffiliation{lbl}{Lawrence Berkeley National Laboratory}
\icmlaffiliation{icsi}{International Computer Science Institute}

\icmlcorrespondingauthor{Pu Ren}{pren@lbl.gov}
\icmlcorrespondingauthor{Yaoqing Yang}{yaoqing.yang@dartmouth.edu}

\icmlkeywords{Scientific Machine Learning, Loss Landscape, Multi-regime Analysis}
\vskip 0.3in
]



\printAffiliationsAndNotice{\icmlEqualContribution} 

\begin{abstract}
Neural networks trained under different hyperparameter settings can fall into distinct training ``regimes,'' with consistent behavior within regimes and qualitative differences across regimes. 
In this paper, we study such multi-regime behavior in scientific machine learning (SciML) models through a \emph{regime-aware diagnostic framework} that jointly analyzes performance, training dynamics, and loss-landscape geometry. 
We identify three key findings: (i) a consistent three-regime structure emerges across many standard SciML models, different constraint enforcements, and various optimizer designs; (ii) optimization effectiveness is regime-specific, with no single method performing well across all regimes; and (iii) SciML models can exhibit fine-grained failure modes that can challenge conventional interpretations of standard loss-landscape metrics. 
Our results provide an approach to establish a unified, task-oblivious perspective on failure modes in SciML and to inform regime-aware guidance for improving robustness. 
We validate these findings across widely-used SciML models, including physics-informed neural networks, neural operators, and neural ordinary differential equations, on benchmarks spanning representative ordinary and partial differential equations. 
\end{abstract}
\section{Introduction}
\label{sec:intro}

Scientific machine learning (SciML) offers the opportunity to combine data-driven ML models with domain-driven physical models. 
Popular SciML model families include: 
(i) physics-informed neural networks (PINNs), which try to enforce physical laws through optimization penalties~\citep{raissi2019unified,karniadakis2021physics}; 
(ii) neural operators (NOs), which try to learn mappings between infinite-dimensional function spaces~\citep{li2020fourier,lu2021learning}; and 
(iii) neural ordinary differential equations (NODEs), which try to parameterize continuous-time dynamics and train via differentiable ODE solvers~\citep{chen2018neural}. 
These approaches have been applied in scientific discovery and engineering design in an effort to accelerate research and improve computational workflows~\citep{karniadakis2021physics,wang2023scientific,azizzadenesheli2024neural}.

Despite the promise of these methods, using them in realistic SciML workflows can be challenging, and recent studies have identified fundamental reasons for these challenges. 
This includes optimization and training difficulties in PINNs~\citep{krishnapriyan2021characterizing}, resolution and aliasing issues in NOs~\citep{sakarvadia2025false}, and discretization and continuity failures in NODEs~\citep{krishnapriyan2023learning}. 
These issues point to a broader ill-posedness: many SciML problems exhibit strong interactions between physical regimes (e.g., stiffness and high-frequency features) and training regimes (e.g., limited supervision and constraint enforcement), leading to behavior that is highly sensitive and often poorly conditioned, and thus difficult to understand from task-level performance alone.

This motivates a diagnostic perspective.
Following prior work on taxonomizing local versus global structure in neural network loss landscapes~\citep{yang2021taxonomizing,zhou2023three}, we adopt a diagnostic approach to analyze and address these challenges.
We provide empirical evidence that, in many cases, the failure modes of SciML models exhibit structured and consistent patterns across commonly-used SciML model families, alongside distinct model-specific behaviors. 
To support this analysis, we develop a \emph{regime-aware diagnostic framework} that systematically maps \emph{when} and \emph{why} these model families exhibit various properties across physical and training regimes. 

Following~\citet{yang2021taxonomizing,zhou2023three}, a \emph{regime} here refers to a region in the configuration space where loss-landscape properties, such as sharpness~\citep{foret2020sharpness,keskar2016large}, connectivity~\citep{garipov2018loss,draxler2018essentially}, and representation similarity~\citep{kornblith2019similarity}, remain qualitatively homogeneous and thus correspond to similar model quality.
Building on this notion, we aim to translate these findings into actionable guidance, including revealing when better-conditioned model designs and optimization methods reduce sensitivity and deliver more stable implicit biases. 
Our approach is motivated in particular by the observation that many SciML exhibits richer and more complex loss landscape structures than typical computer vision (CV) or natural language processing (NLP) models~\citep{li2018visualizing,kwon2021asam,bahri2021sharpness,hooglandloss,chen2025understanding}, and thus more refined diagnostic tools are needed for diagnosing their properties and understanding them. 
As an example, we show empirically, in Section~\ref{sec:sciml_pathological_losslandscape}, that SciML loss landscapes can lack the well-conditioned basins commonly observed in standard CV models, and instead display strong nonconvexity and sensitivity to problem setup. 
This qualitatively-new source of regime structure arises from the interaction among data representation, physical constraints, and optimizations, all of which call for novel diagnostic tools for SciML.

To move beyond ad-hoc hyperparameter tuning and to start to make the above observations operational, we adopt a regime-aware empirical evaluation that systematically varies both \emph{physical regimes} (e.g., parameters and types of partial differential equations (PDEs)) and \emph{training regimes} (e.g., optimizer configurations, constraint handling, and collocation design). 
By jointly analyzing model performance, training dynamics, and loss-landscape structure across these axes, we construct empirical regime maps of model behavior that reveal consistent patterns across SciML models. 
By incorporating problem-dependent structure from the underlying physics and discretization, our framework extends prior work on systematic diagnostic analysis in CV/NLP~\citep{yang2021taxonomizing} to the SciML setting. 
While existing SciML studies have explored related failure phenomena~\citep{krishnapriyan2021characterizing,nncg,krishnapriyan2023learning,sakarvadia2025false}, they are typically limited to a small number of settings and do not provide an operational view of \emph{when} a method is reliable, \emph{why} it fails, and \emph{which interventions} (optimization/training tricks vs.\ better-conditioned model designs) most effectively improve robustness. To summarize, our main contributions are as follows:
\begin{itemize}
\item 
We introduce \emph{a regime-based evaluation taxonomy} that categorizes trained SciML models into three regimes, which are separated by clear boundaries on the training- and test-error plot: Well-Trained, Under-Trained, and Over-Trained. 
This three-regime pattern consistently appears across various model classes, physical systems, and optimization/training methods.

\item 
We develop \emph{a unified evaluation framework} that connects optimizer behavior to loss landscape features and downstream performance. 
Using this framework, we characterize \emph{boundaries of effectiveness} for optimization methods for SciML models, and we show that different approaches are effective in different regimes.

\item 
We identify and analyze previously-uncharacterized and counterintuitive pathological phenomena in SciML models that can cause standard (CV/NLP motivated) performance metrics to misrepresent the underlying loss behavior. 
For example, we observe \emph{deceptive sharpness}, where the Hessian is ill-conditioned such that its trace and leading eigenvalues continue to grow, even as the model becomes increasingly optimized, as indicated by a reduced loss close to zero (Figures~\ref{fig:progressive_sharpening_hessian_plot} and~\ref{fig:regime_hessian_train_loss}(a)). 
We also observe \emph{deceptive flatness}, where the Hessian appears well-behaved while the training loss remains elevated---a clear sign of undertraining (Figures~\ref{fig:regime_hessian_train_loss}(b) and~\ref{fig:initial_flat_losslandscape}). 
We attribute the breakdown of the positive Hessian-loss correlation in SciML to the non-trivial nature of its loss landscape, a fundamental departure from (relatively) well-behaved CV landscapes~\citep{yang2021taxonomizing}. 
We discuss five loss landscape peculiarities that have been associated with training difficulties or failures in CV and NLP models; and we examine which of them do, or do not, explain the specific failures observed in SciML models.

\end{itemize}

Overall, our work provides an empirical regime-based perspective on SciML models that complements existing methodological advances; and, by making failure modes measurable, our framework offers practical guidance for improving robustness and stability in SciML applications. 
For more related work, see Appendix~\ref{app:related_work}; and subsequent appendices contain additional information. Our code is available at \url{https://github.com/leastima/sciml_multi_regime}.

\section{Multi-Regime Analysis Results}
\label{s:results} 

In this section, we present our multi-regime empirical analysis along three complementary axes: model family; physical system; and optimization or training strategy. 
We examine how these factors shape regime geometry, trainability, and generalization. 
We then provide several case studies of loss landscapes observed in SciML tasks.

\subsection{Experimental Setup}
\label{sec:exp_setup}

\paragraph{Datasets and Models.} 
We study representative regime structures across five major SciML models, including PINNs, Fourier NOs (FNOs), Physics-informed NOs (PINOs), NODEs, and Physics-informed NODEs (PINODEs). 
These models are widely-used and conceptually distinct, covering physics-constrained ML methods, operator learning, and temporal dynamics modeling. 
For PINNs, we train on 1D convection, reaction, wave, and reaction-diffusion equations, following the settings in previous work~\citep{krishnapriyan2021characterizing,nncg}. We use 2D Poisson and advection-diffusion systems for FNOs~\citep{subramanian2023towards}, and 2D Darcy flow for PINOs~\citep{li2024physics}; and we use the nonlinear pendulum benchmark in prior work~\citep{krishnapriyan2023learning} for NODEs and PINODEs.

\paragraph{Training and Optimization.} 
We consider a range of optimization and training strategies, including standard first- and second-order methods (Adam~\citep{adam2014method} and L-BFGS~\citep{zhu1997algorithm}), the advanced second-order NysNewton-CG (NNCG) method~\citep{nncg}, hard-constrained optimization via the augmented Lagrangian method (ALM)~\citep{lu2021physics}, curriculum learning (CL)~\citep{krishnapriyan2021characterizing}, and the RoPINN stabilization framework ~\citep{roPINN}. 
These methods play different roles in different algorithms: Adam, L-BFGS, and NNCG modify the parameter-update dynamics; ALM changes the constraint-enforcement formulation; CL alters the training trajectory; and RoPINN aims to improve collocation conditioning through PINN-specific stabilization. 
Thus, we study optimization and training interventions broadly, rather than optimizers in the narrow sense. 
The detailed experimental configurations for each SciML model and benchmark system are summarized in Table~\ref{tab:experiment_setup_summary}. 
Unless otherwise stated, five random seeds are used for PINN and FNO experiments, and three random seeds are used for the remaining experiments. 
The ResNet-18 model~\citep{he2015deepresiduallearningimage} discussed in Section~\ref{sec:sciml_pathological_losslandscape}, which we include for comparison, is trained using stochastic gradient descent~(SGD).

\paragraph{Regime Plots and Boundaries.}
The regime plots we will present (see Figure~\ref{fig:regime_models} for an example) summarize training and test performance across representative SciML models. In each plot, the $x$-axis corresponds to a problem-specific physical parameter, such as the PDE coefficient $\beta$ in the 1D convection equation. Varying this parameter changes the solution structure and target complexity induced by the governing equation, thereby altering training difficulty in a problem-dependent manner. The $y$-axis denotes the amount of supervision, measured by the number of collocation points for PINNs and by the number of training samples for other models. The implementation details of computing Hessian-related quantities, defining regime boundaries, and generating regime plots are described in Appendix~\ref{app:hessian_calculation},~\ref{app:regime_boundary_implement}, and~\ref{app:color_coding}, respectively. 
Specifically, regime boundaries are extracted automatically from the regime diagrams using training- and test-error thresholds. The \textit{generalization boundary} (dashed line) is determined by the test-error threshold $T_{\text{test}}$ and separates regions with successful generalization from those where the model trains but fails to generalize well. The \textit{training boundary} (solid line) is determined by the training-loss threshold $T_{\text{train}}$ and separates trainable regions from those where optimization remains insufficient. To assess robustness, we perturb the percentile thresholds by $\pm 20\%$ and recompute the regime boundaries.

\subsection{Regimes Across SciML Models}
\label{sec:regime_models}

\begin{figure*}[!t]
    \centering
    \begin{tikzpicture}
        \node[anchor=south west, inner sep=0] (img) at (0,0)
        {\includegraphics[width=1.0\textwidth]{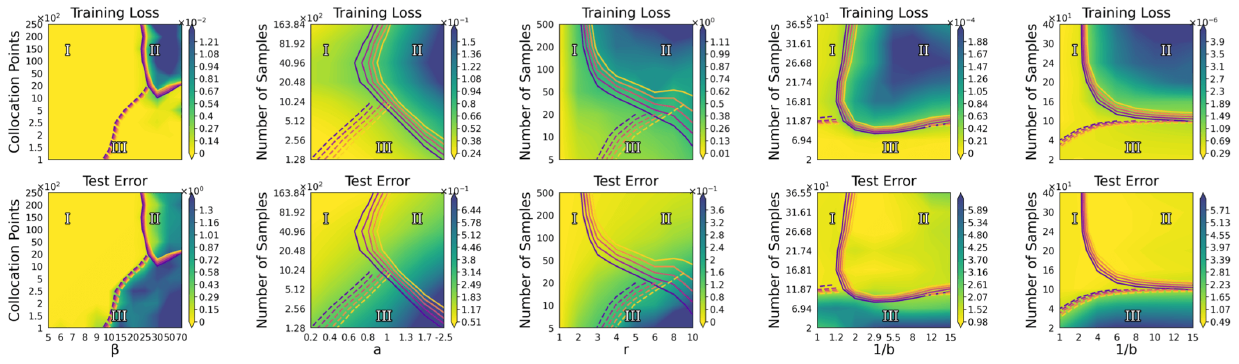}};

        \begin{scope}[x={(img.south east)}, y={(img.north west)}]
            \node at (0.095, -0.04) {\small (\emph{a}) PINN};
            \draw[lightgray, dashed, line width=1pt] (0.195, -0.05) -- (0.195, 1.01);

            \node at (0.30, -0.04) {\small (\emph{b}) FNO};
            \draw[lightgray, dashed, line width=1pt] (0.405, -0.05) -- (0.405, 1.01);
            
            \node at (0.50, -0.04) {\small (\emph{c}) PINO};
            \draw[lightgray, dashed, line width=1pt] (0.605, -0.05) -- (0.605, 1.01);

            \node at (0.71, -0.04) {\small (\emph{d}) NODE};
            \draw[lightgray, dashed, line width=1pt] (0.81, -0.05) -- (0.81, 1.01);

            \node at (0.91, -0.04) {\small (\emph{e}) PINODE};

        \end{scope}
    \end{tikzpicture}

    \caption{
    Representative regime plots across \textbf{varying model families}. Lighter (yellow) and darker (green) colors denote lower and higher training loss/test error, respectively. Across all models, the train-test regime plots consistently separate into three regimes: Regime I (Well-Trained regime with low training and test error); Regime II (Under-Trained regime with high training and test errors); and Regime III (Over-Trained regime with low training loss and high test errors). Colored curves indicate empirically identified regime boundaries under threshold perturbations, showing the robustness and sensitivity of the extracted regime structure.
    }
    \label{fig:regime_models}
\end{figure*}
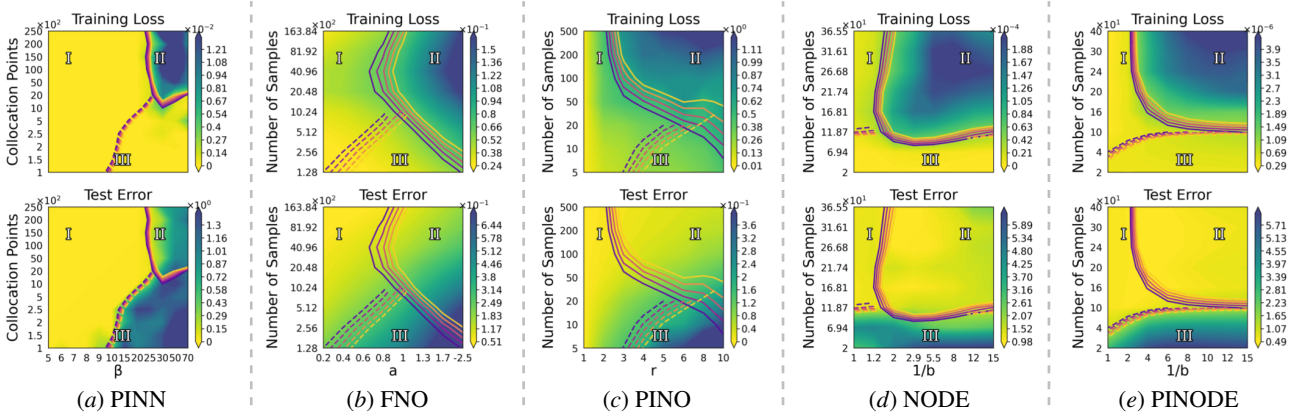

Here, we analyze empirical regimes exhibited by SciML models across different physical and training configurations. 
For each model, we use a benchmark system and its standard training configuration commonly adopted in prior work. 
Specifically, we evaluate PINNs on the 1D convection equation, PINOs on the Darcy flow problems, and PINODEs on the nonlinear pendulum system, using the standard L-BFGS optimizer; and we train FNOs on the 2D advection-diffusion system and NODEs on the pendulum system using Adam. 
We note that these evaluations are not intended as strictly controlled architecture ablations, but instead we use them to provide representative examples for analyzing how different SciML models exhibit distinct regime geometry and transition behavior under realistic training settings. 

Rather than assuming a predefined set of regimes, we identify empirical regimes from the joint behavior of training and test errors over this 2D configuration space. 
Figure~\ref{fig:regime_models} shows that a coarse three-regime structure emerges broadly across diverse SciML models, despite substantial differences in model architectures, training supervisions, and physical systems. These regime labels are \textit{descriptive}: they summarize the observed relationship between training performance, test errors, physical difficulty, and data availability. 
The three empirical regimes are described as follows:
\begin{itemize}
\item 
\textbf{Regime I (Well-Trained).} 
This regime corresponds to configurations where the model achieves low training loss and low test error. 
For the popular SciML models we consider, this regime typically occurs when the physical regime is relatively easy to learn and sufficient training information is available, e.g., with dense collocation points and/or ``nice'' physical parameter settings for PINNs, or with a large number of training samples for FNOs and NODEs. 
In this region, the model can both optimize the training objective and also generalize well to unseen samples.
This regime corresponds to the yellow-colored region marked with ``I''.

\item 
\textbf{Regime II (Under-Trained).} 
This regime is characterized by simultaneously high training and test errors. 
This regime often arises in physically more challenging configurations where the model struggles to optimize the objective, despite having relatively abundant training information. 
In some sense, the dominant failure mode is \textit{optimization difficulty} rather than data scarcity: the model fails to adequately minimize the training objective, and consequently it generalizes poorly. 
This region is labeled ``II''  in the regime figure.

\item 
\textbf{Regime III (Over-Trained).} 
This regime is marked by low training error but high test error. 
This regime appears when training information is limited, such as when PINNs use sparse collocation points or have ``hard'' physical parameters, or when FNOs and NODEs are trained with only a small number of samples. 
This behavior indicates a data-limited failure mode (\textit{degraded generalization}): the model fits the available supervision, but it does not learn a solution that generalizes across the full physical domain or distribution. This region is labeled ``III'' in the regime figure.
\end{itemize}

Although all classes of SciML models exhibit similar regimes and failure modes, the detailed geometry of these regions differs across models. 
Physics-constrained models, particularly PINNs, exhibit sharper and more localized transitions between regimes, indicating stronger sensitivity to physical difficulty and/or training conditions. 
In contrast, NOs and NODEs display smoother and more diffuse transition structure, with less clearly-separated trainability and generalization regions. 
Interestingly, PINOs exhibit intermediate behavior between these extremes, suggesting that incorporating physical constraints into operator-learning architectures partially reintroduces the sharper regime structures observed in PINNs. 
These observations indicate that while coarse failure modes such as trainability and generalization structure appear to be a recurring phenomenon across SciML, the fine-scale geometry of the regimes depends on the balance between data-driven learning, physical constraints, and optimization and training~methods.

\subsection{Regimes Under Varying Physics}
\label{sec:regime_physics}
Here, we consider different physical systems, and we observe that different systems exhibit a similar coarse three-regime structure under a fixed optimization and model setup. 
To illustrate this, we use PINNs as a controlled representative setting (because their collocation-based residual optimization produces sharper regime transitions than other SciML models). 
We use the standard and stable L-BFGS optimizer for training PINNs. The selected PDEs span four representative PDE systems frequently studied in SciML~\citep{krishnapriyan2021characterizing}, including transport-dominated dynamics (convection), reaction-driven dynamics (reaction), oscillatory behavior (wave), and coupled transport-reaction effects (reaction-diffusion). 
Additional regime plots for other SciML models, training strategies, and physical settings are provided in Figures~\ref{fig:regime_physics_alm},~\ref{fig:regime_physics_ropinn}, and~\ref{fig:regime_physics_fno} of Appendix~\ref{app:extened_regime_physics}.

Figure~\ref{fig:regime_physics} shows that a consistent coarse three-regime structure exists across four PDE systems under a fixed PINN and L-BFGS training setup. Despite differences in the underlying physics, all four PDEs exhibit recurring regions associated with stable training and generalization (Regime I), optimization difficulties (Regime II), and degraded generalization (Regime III). This consistency suggests that similar trainability and generalization regimes arise across PDE systems with different physical mechanisms, while PDE-specific structure and parameter-induced difficulty shape the geometry and location of the regime boundaries.
These results demonstrate that while coarse regime organization persists across PDE systems, the fine-scale geometry of the regimes is shaped by the specific physical mechanisms and parameter-induced properties of the underlying equations.

\begin{figure*}[!t]
    \centering
    \begin{tikzpicture}
        \node[anchor=south west, inner sep=0] (img) at (0,0)
        {\includegraphics[width=0.98\textwidth]{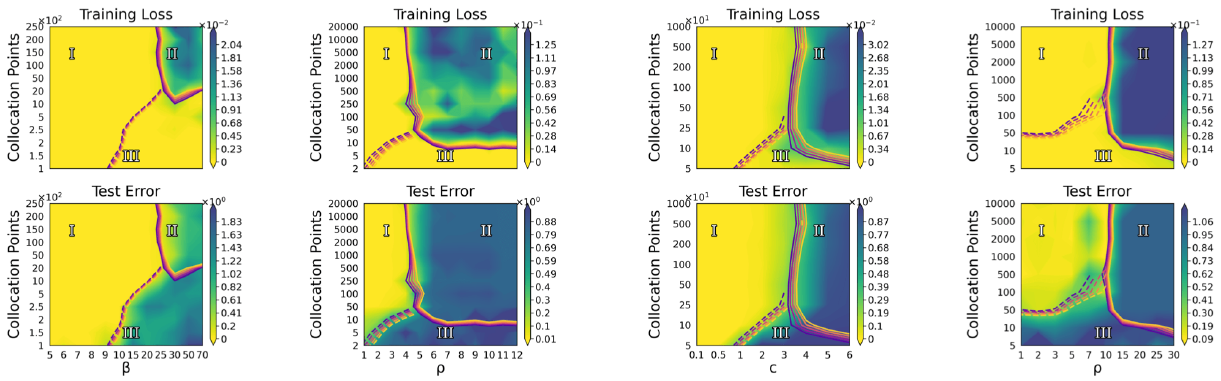}};

        \begin{scope}[x={(img.south east)}, y={(img.north west)}]
            \node at (0.1, -0.04) {\small (\emph{a}) 1D Convection};
            \draw[lightgray, dashed, line width=1pt] (0.235, -0.05) -- (0.235, 1.01);

            \node at (0.365, -0.04) {\small (\emph{b}) 1D Reaction};
            \draw[lightgray, dashed, line width=1pt] (0.5, -0.05) -- (0.5, 1.01);
            
            \node at (0.63, -0.04) {\small (\emph{c}) 1D Wave};
            \draw[lightgray, dashed, line width=1pt] (0.765, -0.05) -- (0.765, 1.01);

            \node at (0.885, -0.04) {\small (\emph{d}) 1D Reaction-Diffusion};
        \end{scope}
    \end{tikzpicture}

    \caption{
    Regime plots across \textbf{varying physical systems} using a fixed PINN architecture trained with the L-BFGS optimizer. The evaluated PDE systems include (\emph{a}) the 1D convection equation with convection coefficient $\beta$, (\emph{b}) the 1D reaction equation with reaction coefficient $\rho$, (\emph{c}) the 1D wave equation with wave speed $c$, and (\emph{d}) the 1D reaction-diffusion equation with reaction coefficient $\rho$.
    }
    \label{fig:regime_physics}
\end{figure*}
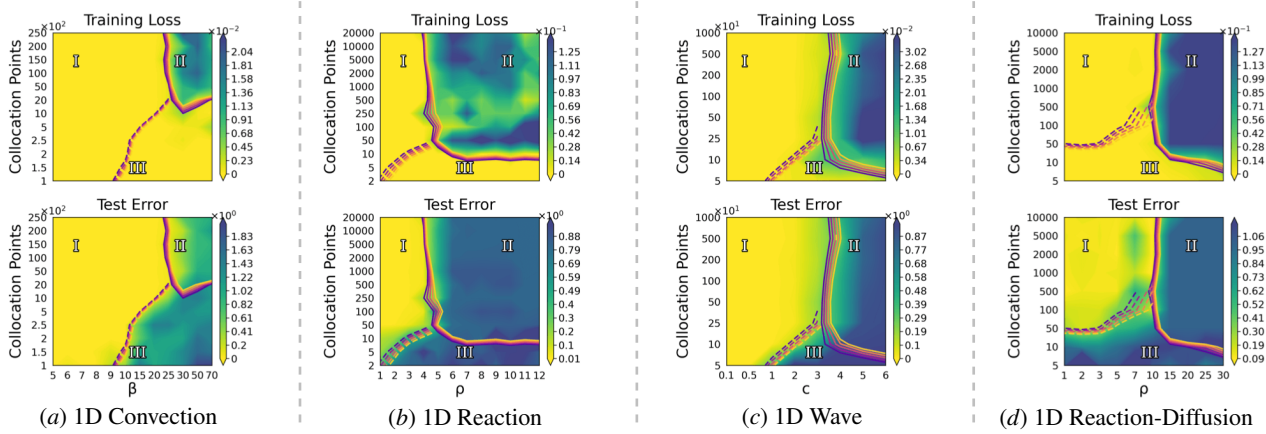

\subsection{Effect of Optimization and Training Strategies}
\label{sec:regime_optimization}
\begin{figure*}[!t]
    \centering
    \begin{tikzpicture}
        \node[anchor=south west, inner sep=0] (img) at (0,0)
        {\includegraphics[width=1.0\textwidth]{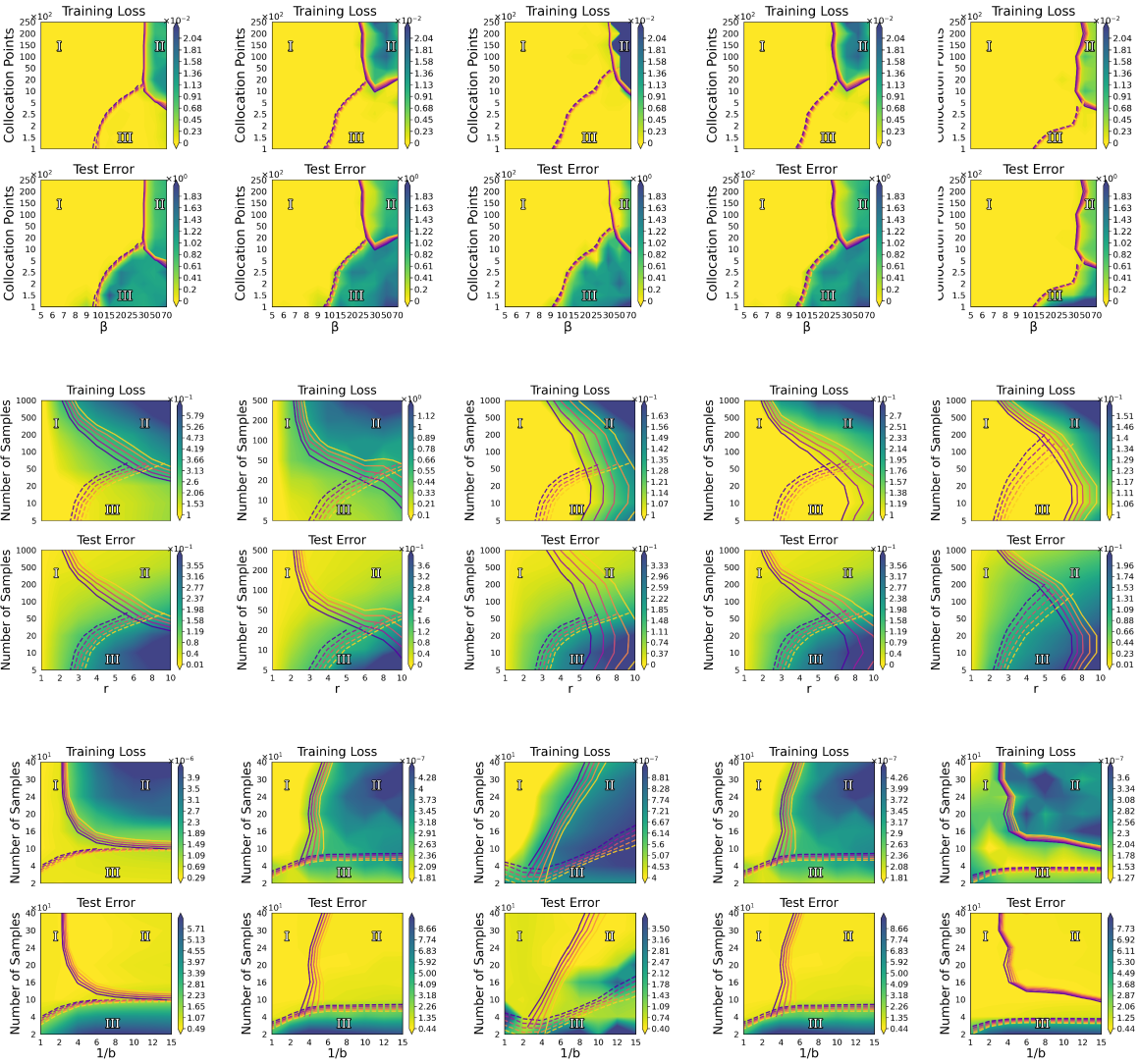}};

        \begin{scope}[x={(img.south east)}, y={(img.north west)}]
            \node at (0.085, 0.67) {\small (\emph{a}) PINN, RoPINN};
            \node at (0.29, 0.67) {\small (\emph{b}) PINN, L-BFGS};
            \node at (0.49, 0.67) {\small (\emph{c}) PINN, ALM};
            \node at (0.7, 0.67) {\small (\emph{d}) PINN, NNCG};
            \node at (0.90, 0.67) {\small (\emph{e}) PINN, CL};
            \draw[lightgray, dashed, line width=1pt] (0.0, 0.655) -- (1.0, 0.655);

            \node at (0.085, 0.33) {\small (\emph{f}) PINO, Adam};
            \node at (0.29, 0.33) {\small (\emph{g}) PINO, L-BFGS};
            \node at (0.49, 0.33) {\small (\emph{h}) PINO, ALM};
            \node at (0.7, 0.33) {\small (\emph{i}) PINO, NNCG};
            \node at (0.90, 0.33) {\small (\emph{j}) PINO, CL};
            \draw[lightgray, dashed, line width=1pt] (0.0, 0.315) -- (1.0, 0.315);

            \node at (0.085, -0.01) {\small (\emph{k}) PINODE, Adam};
            \node at (0.29, -0.01) {\small (\emph{l}) PINODE, L-BFGS};
            \node at (0.49, -0.01) {\small (\emph{m}) PINODE, ALM};
            \node at (0.7, -0.01) {\small (\emph{n}) PINODE, NNCG};
            \node at (0.90, -0.01) {\small (\emph{o}) PINODE, CL};
        
        \end{scope}
    \end{tikzpicture}
    \caption{
    Regime maps for physics-constrained SciML models under \textbf{varying optimization and training strategies}. The first pair of rows presents PINNs on the 1D convection equation using (\emph{a}) RoPINN, (\emph{b}) L-BFGS, (\emph{c}) ALM, (\emph{d}) NNCG, and (\emph{e}) CL. The second pair of rows presents PINOs on the Darcy flow problem using (\emph{f}) Adam, (\emph{g}) L-BFGS, (\emph{h}) ALM, (\emph{i}) NNCG, and (\emph{j}) CL. The third pair of rows presents PINODEs on the nonlinear pendulum system using (\emph{k}) Adam, (\emph{l}) L-BFGS, (\emph{m}) ALM, (\emph{n}) NNCG, and (\emph{o}) CL. 
    }
    \label{fig:regime_optimization}
\end{figure*}
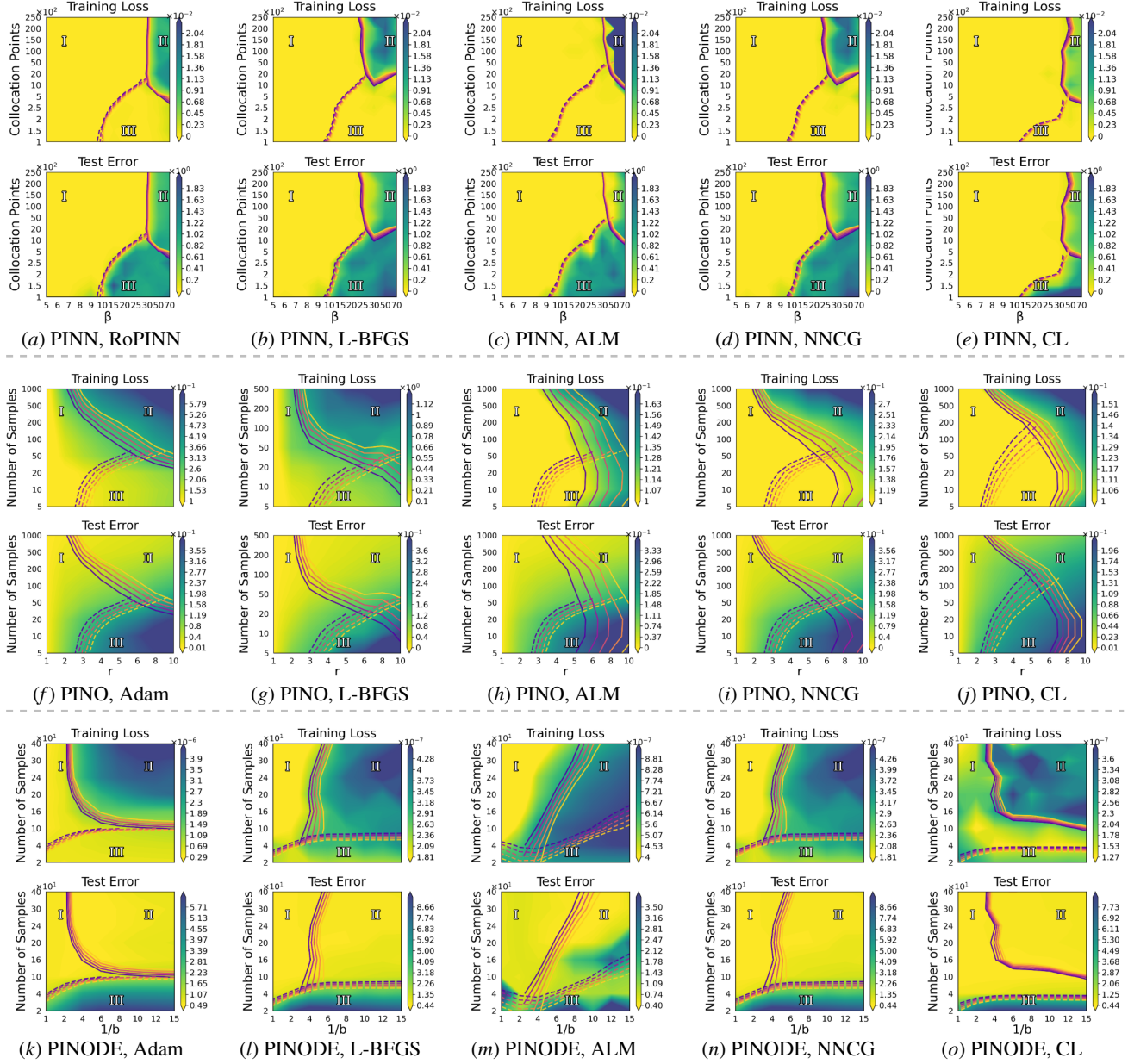

Here, to analyze how optimization and training shape regime geometry, we evaluate three physics-constrained SciML models under a variety of representative optimization and training strategies. 
We consider PINNs on the 1D convection equation, PINOs on the Darcy flow problem, and PINODEs on the nonlinear pendulum system. Figure~\ref{fig:regime_optimization} shows that these interventions primarily reshape the geometry and location of the regime boundaries, rather than eliminating the underlying coarse regime structure. Across all methods, the same coarse regions remain visible, but the detailed shape and extent of these regions vary across interventions.

\paragraph{Effect of Soft and Hard Constraints.}
As shown in Figure~\ref{fig:regime_optimization} (the second and third columns), the comparison between the standard soft-constrained PINN training and the hard-constrained ALM training indicates that the optimization details of constraint handling can shift the trainability boundary in challenging physical regimes. 
For example, the vertical regime transition line in Figure~\ref{fig:regime_optimization}(b) on the training loss plot is at $\beta\approx25$, while the same transition line shifts to $\beta\approx50$ in Figure~\ref{fig:regime_optimization}(c). 
Similar behavior is also observed for PINOs and PINODEs.
That is, ALM expands portions of the trainable region (Regime I) by enforcing the PDE constraint through a hard-constrained augmented Lagrangian formulation, rather than relying only on a soft-constrained penalty method formulation. 
This suggests that improved constraint handling can reduce sensitivity to loss-balancing effects and partially mitigate optimization failure (Regime II). However, ALM does not remove Regime III; instead, it changes its quantitative boundaries.

\paragraph{Effect of CL.}
As shown in Figure~\ref{fig:regime_optimization}(e,j,o), CL improves physics-constrained SciML training in both Regime II (Under-Trained) and Regime III (Over-Trained). In Regime II, CL shifts the Regime I/II boundary toward more challenging physical parameters and lower-data settings, converting some previously under-trained configurations into trainable ones. 
In Regime III, CL reduces test error in parts of the low-data region, but it does not fully remove generalization-limited behavior. For instance, compared with ALM (Figure~\ref{fig:regime_optimization}(c)), CL maintains a comparable training loss boundary to the $\beta=50$; while in Regime III, CL significantly reduces the likelihood of a high generalization gap, outperforming ALM. While ALM requires a fivefold increase in collocation points ($N_f \ge 5000$) to maintain fidelity as $\beta$ rises to 30, CL consistently achieves superior accuracy with as few as $N_f=1000$.  
Moreover, CL produces visibly smoother and less sharply separated transitions than the other strategies. This behavior is consistent with the role of curricula in gradually exposing the model to harder configurations, which changes the optimization trajectory and reduces abrupt changes in trainability. The resulting regime map shows that CL can reshape difficult-regime behavior and modify portions of the boundary.

\paragraph{Effect of RoPINN.}
RoPINN exhibits a distinct regime-shaping effect as a stabilization method rather than a generic optimizer. In Figure~\ref{fig:regime_optimization}(a), it preserves the coarse three-regime structure, but it moves the trainability boundary toward larger convection coefficients $\beta$, indicating improved robustness in more difficult physical regimes with limited collocation points. 
Compared with CL, which smooths regime transitions by modifying the training trajectory, RoPINN produces a sharper boundary shift by improving PINN-specific training stability. However, the generalization-limited regime (Regime III) remains visible, 
especially in low-collocation or high-$\beta$ settings, showing that RoPINN improves trainability but does not fully eliminate generalization failure.

\paragraph{Effect of NNCG.}
The NNCG optimizer, applied here as a post-training fine-tuning stage initialized with the converged L-BFGS solution, demonstrates a distinct performance behavior. While NNCG is designed to use curvature information via a Nyström approximation of the Hessian, our analysis reveals that its stability is sensitive to the physical regime.
Specifically, NNCG presents significant gains in Regime I, corresponding to the low-to-medium coefficient (e.g., $\beta$) regions (the left two-thirds of the phase space). 
However, in the high-difficulty Regime II and Regime III, where $\beta$ is large, the performance of the optimizer is not stable. 
While averaging over multiple random seeds yields a smoother 2D regime plot, individual runs frequently encounter numerical instability or divergence.

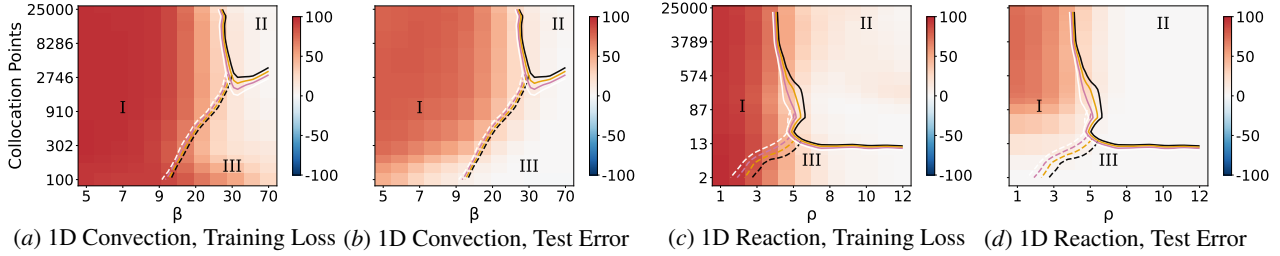
\begin{figure*}[!t]
    \centering
    \begin{tikzpicture}
        \node[anchor=south west, inner sep=0] (img) at (0,0)
        {\includegraphics[width=0.98\textwidth]{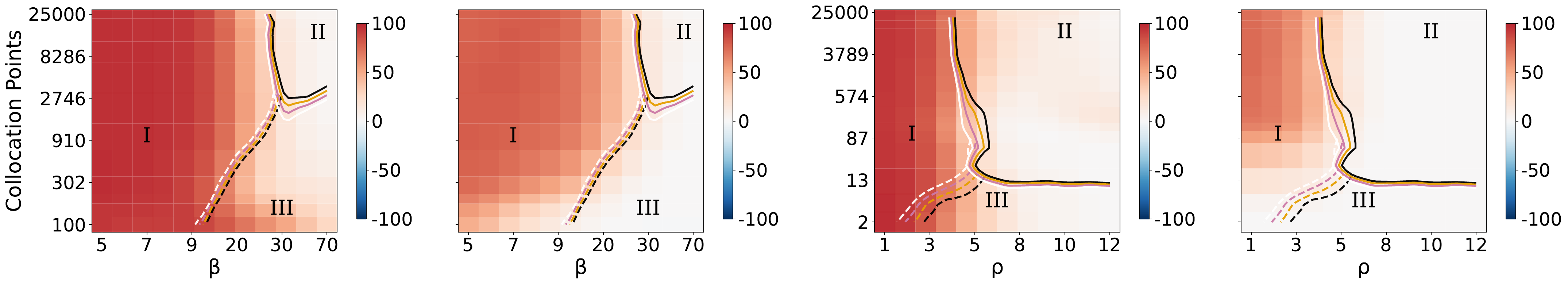}};

        \begin{scope}[x={(img.south east)}, y={(img.north west)}]
            \node at (0.135, -0.05) {\small (\emph{a}) 1D Convection, Training Loss};
            \node at (0.38, -0.05) {\small (\emph{b}) 1D Convection, Test Error};
            \node at (0.64, -0.05) {\small (\emph{c}) 1D Reaction, Training Loss};
            \node at (0.875, -0.05) {\small (\emph{d}) 1D Reaction, Test Error};
        
        \end{scope}
    \end{tikzpicture}
    \caption{
    Relative performance improvement of NNCG over L-BFGS across different physical and data regimes. The heatmaps show the percentile-wise relative improvement of NNCG compared with L-BFGS for training loss and test error, where positive values indicate better performance under NNCG. 
    Panels (\emph{a},\emph{b}) correspond to PINNs trained on the 1D convection equation, while Panels (\emph{c},\emph{d}) present the corresponding results for the 1D reaction equation. 
    In both cases, NNCG presents good fine-tuning gains in Regime~I, indicating improved optimization within already trainable regions.
    }
    \label{fig:nncg_comprehensive_results}
\end{figure*}

This behavior is visualized in Figure~\ref{fig:nncg_comprehensive_results}, which reports the average improvement over three random seeds. In Regime I, NNCG reduces the test error by approximately 50\% in data-limited scenarios and improves training loss by roughly 60\% in well-trained scenarios. Conversely, in the high-$\beta$ parameter space (Regime II), the averaged performance shows minimal improvement, reflecting the difficulty of maintaining a stable Nyström approximation in extremely ill-conditioned landscapes. The success in lower-$\beta$ regimes validates the hypothesis in \citet{nncg} regarding ``under-optimization.'' 
They argue that L-BFGS often terminates prematurely because it cannot find a step size satisfying the strong Wolfe conditions, due to the ``fast spectral decay'' of the Hessian, a claim which our diagnostic analysis validates. 
NNCG circumvents this by using a Nyström preconditioner and a simpler Armijo line search, allowing it to extract further performance in amenable landscapes.

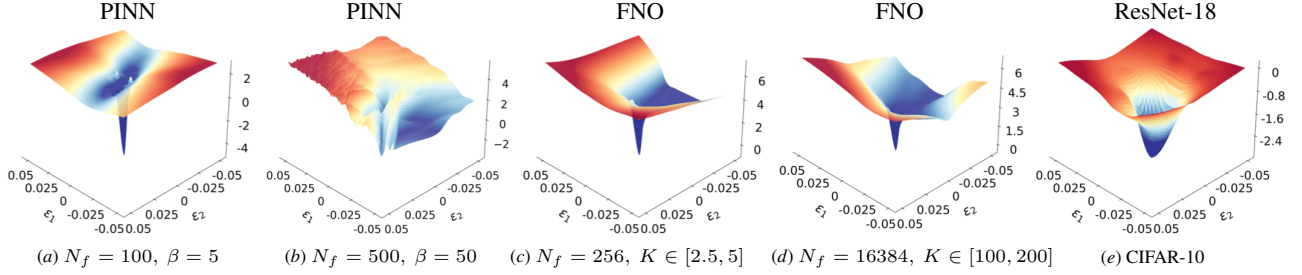
\begin{figure*}[!t]
    \centering
    \begin{tikzpicture}
        \node[anchor=south west, inner sep=0] (img) at (0,0)
        {\includegraphics[
            width=1.0\textwidth
        ]{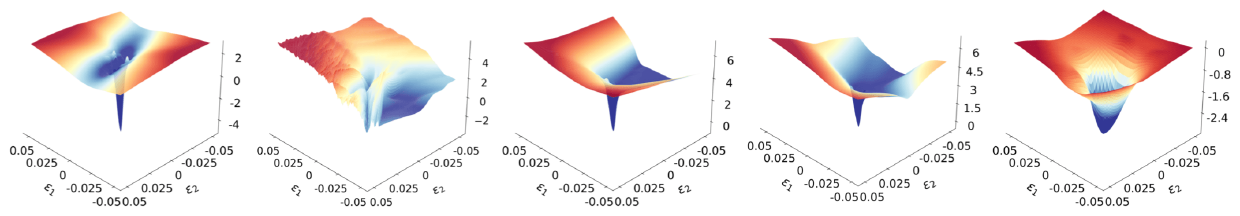}};

        \begin{scope}[x={(img.south east)}, y={(img.north west)}]
            \node at (0.1, 1.03) {\small PINN};
            \node at (0.1, -0.06) {\scriptsize (\emph{a}) $N_f=100,\ \beta=5$};

            \node at (0.29, 1.03) {\small PINN};
            \node at (0.295, -0.06) {\scriptsize (\emph{b}) $N_f=500,\ \beta=50$};

            
            \node at (0.495, 1.03) {\small FNO};
            \node at (0.485, -0.06) {\scriptsize (\emph{c}) $N_f=256,\ K \in [2.5,5]$};
            

            \node at (0.695, 1.03) {\small FNO};
            \node at (0.705, -0.06) {\scriptsize (\emph{d}) $N_f=16384,\ K \in [100,200]$};


            \node at (0.9, 1.03) {\small ResNet-18};
            \node at (0.89, -0.06) {\scriptsize (\emph{e}) CIFAR-10};
            
        \end{scope}
    \end{tikzpicture}

    \caption{
    Comparison of 3D loss landscapes across PINN, FNO, and ResNet models. Panels (\emph{a},\emph{b}) and (\emph{c},\emph{d}) visualize the non-convex landscapes of a PINN (trained on 1D Convection with different numbers of collocation points $N_f$ and PDE coefficients $\beta$) and an FNO (trained on 2D Poisson with varying numbers of training samples $N_f$ and frequency modes $K$), respectively. They are characterized by sharp local minima and chaotic ridges. In contrast, Panel (\emph{e}) displays the smooth, convex basin of a standard ResNet-18.
    }
    \label{fig:3d_losslandscape_cv_sciml}
\end{figure*}

\subsection{Case Studies of SciML Landscapes}
\label{sec:sciml_pathological_losslandscape}

Here, we present case studies of loss landscapes in SciML. We first examine their \emph{topological characteristics}, showing that SciML landscapes often contain sharper, more disconnected, and more anisotropic structures than those observed in standard CV settings. We then analyze \emph{nonstandard landscape phenomena}, where popular and intuitive landscape metrics can give misleading signals about trainability and generalization. Together, these case studies provide a finer-scale characterization beyond Hessian-based analysis~\citep{krishnapriyan2021characterizing}, and they reveal distinct optimization pathologies in commonly-used SciML models.

\subsubsection{SciML Topological Characteristics}
\label{sec:sciml_topological_char}

\paragraph{More Rugged Landscapes in SciML.} 
Figure~\ref{fig:3d_losslandscape_cv_sciml} compares the 3D loss landscapes of SciML models (PINNs and FNO) with that of a well-trained CV model (ResNet-18). 
Consistent with prior analyses~\citep{li2018visualizing}, the CV loss surface forms a wide, gently curved basin with relatively smooth curvature. In contrast, the PINN landscapes exhibit distinct, regime-dependent patterns. When the PDE coefficient is small, indicating an easier-to-solve problem, the landscape features a large flat region of high loss surrounding a deep, narrow valley (Figure~\ref{fig:3d_losslandscape_cv_sciml}(a)). However, as the parameter increases, corresponding to a progressively more challenging regime, the landscape transitions into a highly irregular and much more rugged state. While increasing the number of collocation points offers slight mitigation (Figure~\ref{fig:3d_losslandscape_cv_sciml}(b)), the topology remains more rugged than the well-behaved CV counterparts. 
This is consistent with observations in prior work \citep{krishnapriyan2021characterizing}. 
Furthermore, this phenomenon extends to operator learning models; both PINN and FNO exhibit sharp minima, whereas ResNet converges to a smoother minimum, underscoring the intrinsic optimization challenges of these widely used SciML models.

\begin{figure*}[ht!]
  \centering
    \begin{tikzpicture}
        \node[anchor=south west, inner sep=0] (img) at (0,0)
        {\includegraphics[width=1.0\textwidth]{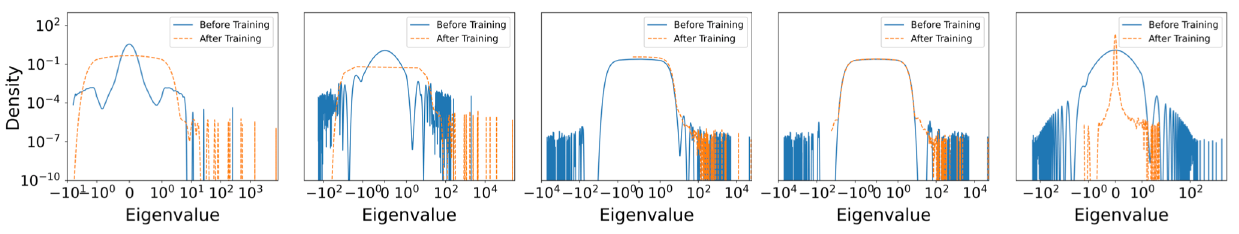}};

        \begin{scope}[x={(img.south east)}, y={(img.north west)}]
            \node at (0.13, 1.03) {\small PINN};
            \node at (0.128, -0.06) {\scriptsize (\emph{a}) $N_f=100,\ \beta=5$};

            \node at (0.33, 1.03) {\small PINN};
            \node at (0.325, -0.06) {\scriptsize (\emph{b}) $N_f=500,\ \beta=50$};

            
            \node at (0.52, 1.03) {\small FNO};
            \node at (0.515, -0.06) {\scriptsize (\emph{c}) $N_f=256,\ K \in [2.5,5]$};
            

            \node at (0.71, 1.03) {\small FNO};
            \node at (0.72, -0.06) {\scriptsize (\emph{d}) $N_f=16384,\ K \in [100,200]$};


            \node at (0.9, 1.03) {\small ResNet-18};
            \node at (0.9, -0.06) {\scriptsize (\emph{e}) CIFAR-10};
            
        \end{scope}
    \end{tikzpicture}

    \caption{
    Comparison of Hessian eigenspectra across PINN, FNO, and ResNet models. Each panel shows the empirical spectral density of the Hessian before and after training. Panels (\emph{a},\emph{b}) show PINNs trained on the 1D Convection under different physical and data regimes, varying the number of collocation points $N_f$ and the convection coefficient $\beta$. Panels (\emph{c},\emph{d}) show FNOs trained on the 2D Poisson under different training-sample sizes $N_f$ and coefficient ranges $K$. Panel (\emph{e}) shows ResNet-18 trained on CIFAR-10 as a CV reference. 
    \vspace{-2mm}
    }
  \label{fig:hessian_comparison_combined}
\end{figure*}
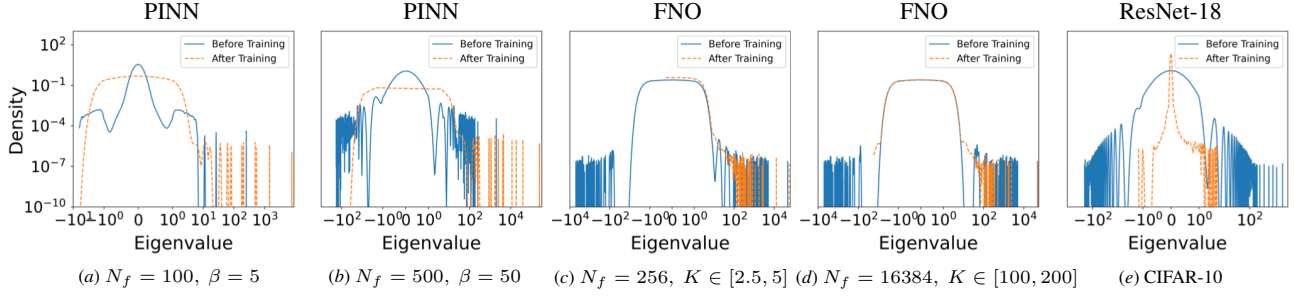

\paragraph{Lack of Connectivity and Existence of Large Hessian Eigenvalues.} 
In Figure~\ref{fig:hessian_comparison_combined}(e), we present the estimated Hessian spectral density of ResNet-18 model.
Two characteristics are worth noting: the largest outlier eigenvalue is relatively small (around 10); and there is a density peak near zero. This abundance of near-zero flat directions has been observed in prior studies on the Hessian spectrum~\citep{sagun2017empirical}. Consequently, the zero-loss manifold of the loss landscape in CV classification models is often conceptualized as being highly connected and is supported by \emph{mode connectivity}~\citep{draxler2018essentially,garipov2018loss}. 
However, for the popular SciML models analyzed in this paper, the eigenvalues exhibit a markedly different distribution. As shown in Figures~\ref{fig:hessian_comparison_combined}(a,b), which present the estimated spectral densities of PINNs, we observe large outlier eigenvalues (exceeding $10^3$ or even $10^4$) and the absence of a zero-eigenvalue peak, a phenomenon that becomes even more significant after training.
This spectral density suggests that the optimization trajectories are often trapped in spurious minima that hinder further descent, preventing the model from recovering the correct physical behavior. In such cases, the large outlier eigenvalues indicate that the neighborhood around the loss minimum is extremely sharp, implying that more effective optimization methods are required for SciML models to escape local minima and converge toward regions near the global minimum.

\paragraph{Landscapes Decouple Hessian–loss Correlation.}  
The distinct Hessian spectral densities observed in these models suggest that the conventional Hessian-loss relationship, as popularly conceptualized within ML, does not necessarily port to SciML models.  
In standard CV tasks, minimizing the loss (a global property) typically correlates with a reduction in curvature (a local property); as the model approaches the solution, the Hessian eigenvalues tend to diminish, and the distribution shifts towards zero (Figure~\ref{fig:hessian_comparison_combined}(e)). However, for many SciML models, we observe a decoupling of this correlation. 
Specifically, while the training loss decreases, the leading Hessian eigenvalues do not vanish, but instead they remain large or can even increase. We hypothesize that this phenomenon is related to \emph{increasing sharpening}~\cite{damian2022self}, where sharpness continues to grow as models approach the bottom of the basin. This results in a loss landscape characterized by a thin, funnel-shaped structure near the minimum, as opposed to the wide, flat basins of CV models. Consequently, a low training loss in PINNs does not guarantee a robust solution, as the increasing PDE coefficients can significantly sharpen the landscape, causing Hessian-based metrics to diverge from the loss trajectory.

\subsubsection{Beyond Standard Landscape Intuitions}
\label{sec:possible_failure}

The Hessian-loss relationship discussed in Section~\ref{sec:sciml_topological_char} shows that standard loss landscape intuitions from conventional deep learning can fail to characterize SciML model behavior. In particular, a low training loss does not necessarily correspond to a low curvature, nor does a high loss imply high curvature. We discuss two possible types of ``pathological'' loss landscapes arising from this: 
(i) the solution is hidden within sharp and narrow basins (\textit{deceptive sharpness}); and (ii) flatness masks a lack of learning signal (\textit{deceptive flatness}). 
We also examine three alternative explanations commonly studied in deep learning that could plausibly account for SciML optimization difficulty: landscape jumps between basins; landscape barriers; and landscape aging. Our results indicate that these mechanisms do not fully explain the observed SciML behavior.

\begin{figure}[hthb]
    \centering
    \begin{subfigure}{0.4\textwidth}
        \centering
        \includegraphics[width=\linewidth]{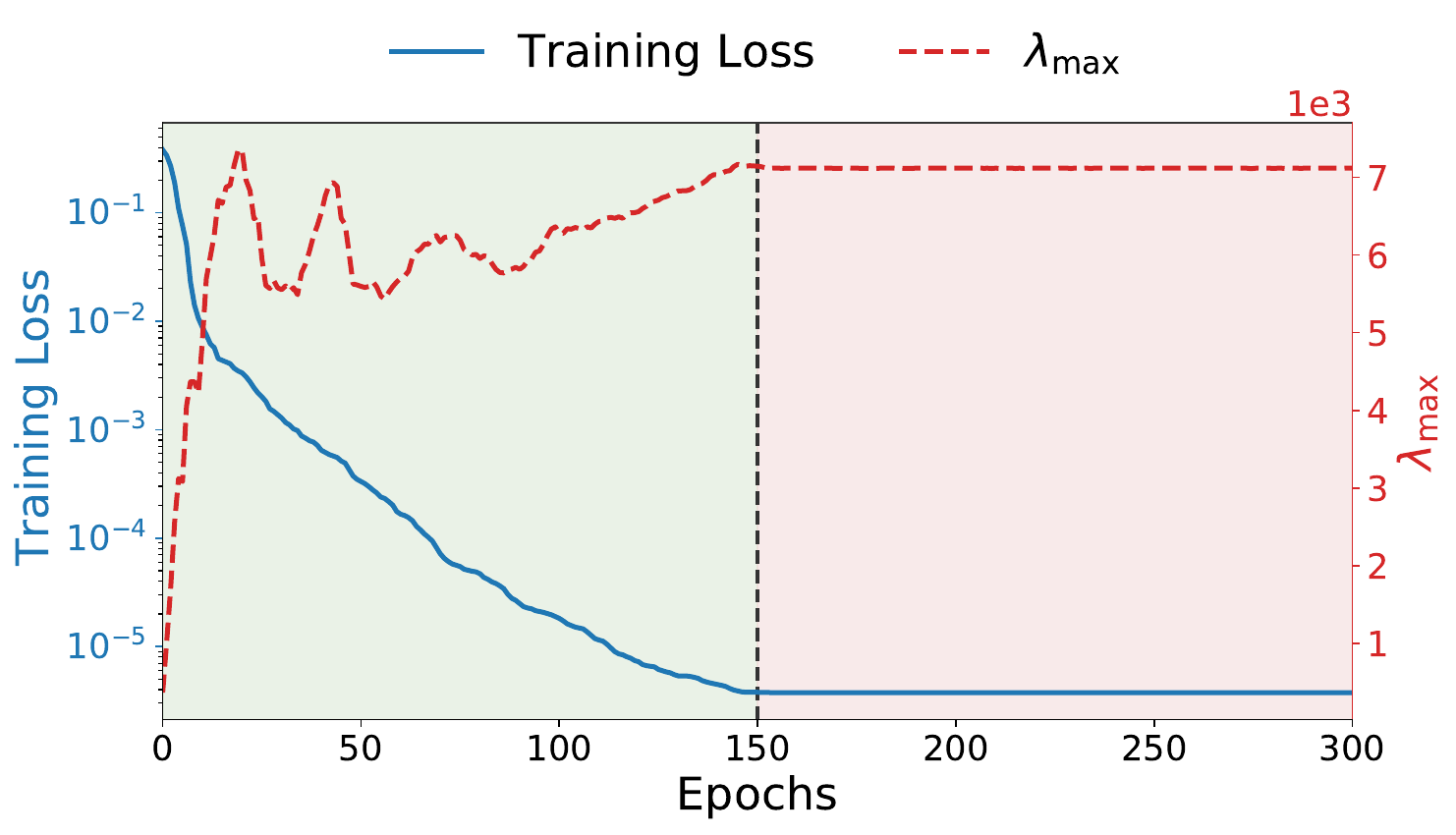}
    \end{subfigure}
    \caption{Increasing sharpening dynamics on the 1D convection problem ($N_f=15000, \beta=5.0$) trained with the L-BFGS optimizer with five random seeds. The plots illustrate the correlation between curvature and loss: $\lambda_{\max}$ rises sharply during the initial phase, coinciding with the rapid decay in training loss.\vspace{-3mm}}
    \label{fig:progressive_sharpening_hessian_plot} 
\end{figure}

\paragraph{Deceptive Sharpness.}
High Hessian eigenvalues do not necessarily indicate poor training. 
Rather, they can indicate the entrance to more optimal and well-trained regions, which are often hypothesized to be reached through narrow and sharp bottleneck paths~\citep{golatkar2019time}. 
This geometry not only poses a significant challenge for optimizers, but it also explains the sudden loss drop observed once these bottlenecks are successfully traversed~\citep{cohen2021gradient,damian2022self}. 
We visualize this phenomenon in Figure~\ref{fig:progressive_sharpening_hessian_plot}. The training dynamics exhibit a phase of \textit{Increasing Sharpening} (indicated by the green background), while the maximum Hessian eigenvalue $\lambda_{\max}$ undergoes a dramatic ascent. This sharp increase in curvature does not hinder optimization. 
On the contrary, it perfectly coincides with the most rapid decrease in training loss. This suggests that the optimizer is successfully navigating into a sharp but highly accurate valley. Following this, the model enters a \textit{Stable Regime} (pink background), where both the loss and the sharpness stabilize. More examples can be found in Figure~\ref{fig:extend_increasing_sharpness} of Appendix~\ref{app:extend_deceptive_sharpness}.

This counterintuitive observation that sharper minima can yield lower training errors is not limited to the temporal dynamics, but it also manifests across different problem settings. As corroborated by our additional analysis on the 2D Poisson problem (see Figure~\ref{fig:regime_hessian_train_loss}(a) of Appendix~\ref{app:extend_deceptive_sharpness}), we observe an anomalous distribution where well-trained regions (Regime I) are characterized by high Hessian eigenvalues. In contrast, regions with lower Hessian values (Regime II) correspond to significantly higher training errors, indicating that those ``flatter'' areas are likely distinct failure modes or deceptive plateaus rather than generalizable minima.

\paragraph{Deceptive Flatness and Vanishing Descent Directions.}
While flat minima are typically desired for their generalization properties, our analysis reveals a counter-phenomenon: \textit{deceptive flatness}. Detailed analysis of the Hessian spectrum and phase plots is provided in Appendix~\ref{app:extend_deceptive_flatness}, demonstrating that flat regions in SciML often occur at high training-loss values.
Unlike the wide, low-loss basins associated with robust solutions, these high-loss plateaus correspond to a lack of informative gradients. Therefore, the optimizer receives negligible directional signals, causing it to stall.
This deceptive structure is particularly pronounced in regimes with larger, more difficult PDE coefficients (e.g., $\rho \ge 15$). As detailed in Figure~\ref{fig:regime_hessian_train_loss}(b) of Appendix~\ref{app:extend_deceptive_flatness}, models in these settings frequently become trapped in regions characterized by small Hessian eigenvalues yet significantly elevated training losses (Regime II/III in the phase plots). 
Theoretically, this observation resonates with the sigmoidal learning dynamics analyzed by~\citet{saxe2014exact}, where models must navigate long plateaus. However, the persistence of the plateau in our experiments suggests a more fundamental issue rooted in the sharpness-diversity tradeoff~\cite{lu2024sharpness, gong2025makes}. Specifically, the observed flatness does not stem from a wide basin of attraction, but rather from \textit{reduced information content} in the gradients. This lack of driving force creates a ``blind'' landscape where the optimizer, deprived of directional guidance, fails to trigger the effective descent phase.

\paragraph{Loss Landscape Jumps Between Basins.}
We first test whether SciML optimization difficulty can be explained by repeated jumps between disconnected basins. However, our observation suggests the opposite: PINNs, especially in harder physical regimes, exhibit fewer basin transitions than ResNet-18. Thus, SciML failures are less likely driven by excessive basin jumping and more likely caused by degraded, poorly conditioned landscape geometry. See Appendix~\ref{app:unlikely_failure_model_1} for details.

\vspace{-1mm}
\paragraph{Loss Landscape Barriers.}
We next test whether SciML optimization difficulty can be explained by the existence of high-loss barriers separating coarse and fine solutions. 
However, our analysis shows that the primary difficulty in PINNs is not barrier crossing, but the mismatch between low training loss and physically correct solutions (Appendix~\ref{app:unlikely_failure_model_2}).

\vspace{-1mm}
\paragraph{Loss Landscape Aging.}
Finally, we test whether SciML optimization difficulty can be explained by loss landscape aging~\cite{baity2018aging_reference}, where optimization slows after entering a wide and flat minimum. Our results demonstrate that this mechanism does not match the observed SciML behavior: PINN landscapes retain many nonzero Hessian directions and remain relatively sharp even near convergence. Thus, the difficulty is unlikely to be primarily caused by excessive flatness or diffusive late-stage dynamics. See Appendix~\ref{app:unlikely_failure_model_3} for details.

\section{Conclusion}
\label{sec:conclusion}
\vspace{-2mm}
We present a regime-aware empirical and diagnostic analysis of SciML models across ML model families, PDE systems, and optimization and training strategies. 
A recurring coarse three-regime structure is observed associated with \textit{stable training} (Regime I), \textit{optimization difficulty} (Regime II), and \textit{degraded generalization} (Regime III). 
Our results show that training and optimization strategies mainly reshape regime boundaries rather than eliminating the underlying structure. 
Loss landscape case studies further reveal SciML-specific geometries that challenge standard interpretations of flatness, sharpness, and optimization stability. 
More generally, these findings provide a foundation for regime-aware diagnostics and more robust SciML training~methods.




\clearpage
\section*{Acknowledgements}
YY would like to acknowledge the support by the U.S. Department of Energy (Grant number: DE-SC0025584) and the Defense Advanced Research Projects Agency (Grant number: HR00112520011). MWM would like to acknowledge DARPA, DOE, NSF, ONR, and the DOE SciGPT grant. We would also like to acknowledge NERSC DOE Mission Science Allocation for ERCAP request ERCAP0035003.
\section*{Impact Statement}
This paper presents research aimed at advancing the fields of Machine Learning and Scientific Computing, particularly in diagnosing and optimizing Scientific Machine Learning models through loss landscape and regime analysis. While our work has various potential societal implications, we do not find it necessary to highlight any specific ones here.
\bibliography{refs}

@inproceedings{nakkiran2019deep,
  author       = {Preetum Nakkiran and
                  Gal Kaplun and
                  Yamini Bansal and
                  Tristan Yang and
                  Boaz Barak and
                  Ilya Sutskever},
  title        = {Deep Double Descent: Where Bigger Models and More Data Hurt},
  booktitle    = {{ICLR}},
  publisher    = {OpenReview.net},
  year         = {2020}
}

@article{wei2022emergent,
  author       = {Jason Wei and
                  Yi Tay and
                  Rishi Bommasani and
                  Colin Raffel and
                  Barret Zoph and
                  Sebastian Borgeaud and
                  Dani Yogatama and
                  Maarten Bosma and
                  Denny Zhou and
                  Donald Metzler and
                  Ed H. Chi and
                  Tatsunori Hashimoto and
                  Oriol Vinyals and
                  Percy Liang and
                  Jeff Dean and
                  William Fedus},
  title        = {Emergent Abilities of Large Language Models},
  journal      = {Trans. Mach. Learn. Res.},
  volume       = {2022},
  year         = {2022}
}

@article{schaeffer2024emergent,
  title={Are emergent abilities of large language models a mirage?},
  author={Schaeffer, Rylan and Miranda, Brando and Koyejo, Sanmi},
  journal={Advances in Neural Information Processing Systems},
  volume={36},
  year={2024}
}

@inproceedings{yang2021taxonomizing,
  title = {Taxonomizing Local versus Global Structure in Neural Network Loss Landscapes},
  booktitle = {Advances in {{Neural Information Processing Systems}}},
  author = {Yang, Yaoqing and Hodgkinson, Liam and Theisen, Ryan and Zou, Joe and Gonzalez, Joseph E and Ramchandran, Kannan and Mahoney, Michael W},
  year = {2021},
  volume = {34},
  pages = {18722--18733},
  publisher = {Curran Associates, Inc.}
}

@inproceedings{zhou2023three,
  title={A three-regime model of network pruning},
  author={Zhou, Yefan and Yang, Yaoqing and Chang, Arin and Mahoney, Michael W},
  booktitle={International Conference on Machine Learning},
  pages={42790--42809},
  year={2023},
  organization={PMLR}
}

@inproceedings{kwon2021asam,
  title={Asam: Adaptive sharpness-aware minimization for scale-invariant learning of deep neural networks},
  author={Kwon, Jungmin and Kim, Jeongseop and Park, Hyunseo and Choi, In Kwon},
  booktitle={International Conference on Machine Learning},
  pages={5905--5914},
  year={2021},
  organization={PMLR}
}

@inproceedings{andriushchenko2023modern,
  author       = {Maksym Andriushchenko and
                  Francesco Croce and
                  Maximilian M{\"{u}}ller and
                  Matthias Hein and
                  Nicolas Flammarion},
  title        = {A Modern Look at the Relationship between Sharpness and Generalization},
  booktitle    = {{ICML}},
  series       = {Proceedings of Machine Learning Research},
  pages        = {840--902},
  publisher    = {{PMLR}},
  year         = {2023}
}

@inproceedings{theisenWhenAreEnsembles2023,
  author       = {Ryan Theisen and
                  Hyunsuk Kim and
                  Yaoqing Yang and
                  Liam Hodgkinson and
                  Michael W. Mahoney},
  title        = {When are ensembles really effective?},
  booktitle    = {NeurIPS},
  year         = {2023}
}

@inproceedings{hao2023pinnacle,
  author       = {Zhongkai Hao and
                  Jiachen Yao and
                  Chang Su and
                  Hang Su and
                  Ziao Wang and
                  Fanzhi Lu and
                  Zeyu Xia and
                  Yichi Zhang and
                  Songming Liu and
                  Lu Lu and
                  Jun Zhu},
  title        = {PINNacle: {A} Comprehensive Benchmark of Physics-Informed Neural Networks
                  for Solving PDEs},
  booktitle    = {NeurIPS},
  year         = {2024}
}

@article{raissi2019unified,
    title = {Physics-informed neural networks: A deep learning framework for solving forward and inverse problems involving nonlinear partial differential equations},
    journal = {Journal of Computational Physics},
    volume = {378},
    pages = {686-707},
    year = {2019},
    author = {M. Raissi and P. Perdikaris and G.E. Karniadakis}
}

@inproceedings{krishnapriyan2021characterizing,
 author = {Krishnapriyan, Aditi and Gholami, Amir and Zhe, Shandian and Kirby, Robert and Mahoney, Michael W},
 booktitle = {Advances in Neural Information Processing Systems},
 title = {Characterizing possible failure modes in physics-informed neural networks},
 year = {2021}
}

@article{li2020neural,
  title={Neural operator: Graph kernel network for partial differential equations},
  author={Li, Zongyi and Kovachki, Nikola and Azizzadenesheli, Kamyar and Liu, Burigede and Bhattacharya, Kaushik and Stuart, Andrew and Anandkumar, Anima},
  journal={arXiv preprint arXiv:2003.03485},
  year={2020}
}

@inproceedings{li2020fourier,
  author       = {Zongyi Li and
                  Nikola Borislavov Kovachki and
                  Kamyar Azizzadenesheli and
                  Burigede Liu and
                  Kaushik Bhattacharya and
                  Andrew M. Stuart and
                  Anima Anandkumar},
  title        = {Fourier Neural Operator for Parametric Partial Differential Equations},
  booktitle    = {{ICLR}},
  publisher    = {OpenReview.net},
  year         = {2021}
}

@article{subramanian2023towards,
  title={Towards foundation models for scientific machine learning: Characterizing scaling and transfer behavior},
  author={Subramanian, Shashank and Harrington, Peter and Keutzer, Kurt and Bhimji, Wahid and Morozov, Dmitriy and Mahoney, Michael W and Gholami, Amir},
  journal={Advances in Neural Information Processing Systems},
  volume={36},
  pages={71242--71262},
  year={2023}
}

@article{li2018visualizing,
  title={Visualizing the loss landscape of neural nets},
  author={Li, Hao and Xu, Zheng and Taylor, Gavin and Studer, Christoph and Goldstein, Tom},
  journal={Advances in neural information processing systems},
  volume={31},
  year={2018}
}

@article{lu2021physics,
  title={Physics-informed neural networks with hard constraints for inverse design},
  author={Lu, Lu and Pestourie, Raphael and Yao, Wenjie and Wang, Zhicheng and Verdugo, Francesc and Johnson, Steven G},
  journal={SIAM Journal on Scientific Computing},
  volume={43},
  number={6},
  pages={B1105--B1132},
  year={2021},
  publisher={SIAM}
}

@article{yao2018large,
  title={Large batch size training of neural networks with adversarial training and second-order information},
  author={Yao, Zhewei and Gholami, Amir and Arfeen, Daiyaan and Liaw, Richard and Gonzalez, Joseph and Keutzer, Kurt and Mahoney, Michael},
  journal={arXiv preprint arXiv:1810.01021},
  year={2018}
}

@inproceedings{yao2020pyhessian,
  title={Pyhessian: Neural networks through the lens of the hessian},
  author={Yao, Zhewei and Gholami, Amir and Keutzer, Kurt and Mahoney, Michael W},
  booktitle={2020 IEEE international conference on big data (Big data)},
  pages={581--590},
  year={2020},
  organization={IEEE}
}

@inproceedings{foret2020sharpness,
  author       = {Pierre Foret and
                  Ariel Kleiner and
                  Hossein Mobahi and
                  Behnam Neyshabur},
  title        = {Sharpness-aware Minimization for Efficiently Improving Generalization},
  booktitle    = {{ICLR}},
  publisher    = {OpenReview.net},
  year         = {2021}
}

@article{hooglandloss,
  title={Loss Landscape Degeneracy and Stagewise Development in Transformers},
  author={Hoogland, Jesse and Wang, George and Farrugia-Roberts, Matthew and Carroll, Liam and Wei, Susan and Murfet, Daniel},
  journal={Transactions on Machine Learning Research},
  year={2024}
}

@article{xie2024losslens,
  title={LossLens: Diagnostics for Machine Learning Through Loss Landscape Visual Analytics},
  author={Xie, Tiankai and Chen, Jiaqing and Yang, Yaoqing and Geniesse, Caleb and Shi, Ge and Chaudhari, Ajinkya and Cava, John Kevin and Mahoney, Michael W and Perciano, Talita and Weber, Gunther H and others},
  journal={IEEE Computer Graphics and Applications},
  year={2024},
  publisher={IEEE}
}

@article{keskar2016large,
  title={On large-batch training for deep learning: Generalization gap and sharp minima},
  author={Keskar, Nitish Shirish and Mudigere, Dheevatsa and Nocedal, Jorge and Smelyanskiy, Mikhail and Tang, Ping Tak Peter},
  journal={arXiv preprint arXiv:1609.04836},
  year={2016}
}

@article{garipov2018loss,
  title={Loss surfaces, mode connectivity, and fast ensembling of dnns},
  author={Garipov, Timur and Izmailov, Pavel and Podoprikhin, Dmitrii and Vetrov, Dmitry P and Wilson, Andrew G},
  journal={Advances in neural information processing systems},
  volume={31},
  year={2018}
}

@inproceedings{draxler2018essentially,
  title={Essentially no barriers in neural network energy landscape},
  author={Draxler, Felix and Veschgini, Kambis and Salmhofer, Manfred and Hamprecht, Fred},
  booktitle={International conference on machine learning},
  pages={1309--1318},
  year={2018},
  organization={PMLR}
}

@inproceedings{kornblith2019similarity,
  title={Similarity of neural network representations revisited},
  author={Kornblith, Simon and Norouzi, Mohammad and Lee, Honglak and Hinton, Geoffrey},
  booktitle={International conference on machine learning},
  pages={3519--3529},
  year={2019},
  organization={PMlR}
}

@inproceedings{nncg,
  author       = {Pratik Rathore and
                  Weimu Lei and
                  Zachary Frangella and
                  Lu Lu and
                  Madeleine Udell},
  title        = {Challenges in Training PINNs: {A} Loss Landscape Perspective},
  booktitle    = {{ICML}},
  series       = {Proceedings of Machine Learning Research},
  pages        = {42159--42191},
  publisher    = {{PMLR} / OpenReview.net},
  year         = {2024}
}

@article{chen2025understanding,
  title={Unveiling the Basin-Like Loss Landscape in Large Language Models},
  author={Chen, Huanran and Dong, Yinpeng and Wei, Zeming and Huang, Yao and Zhang, Yichi and Su, Hang and Zhu, Jun},
  journal={arXiv preprint arXiv:2505.17646},
  year={2025}
}

@inproceedings{bahri2021sharpness,
  author       = {Dara Bahri and
                  Hossein Mobahi and
                  Yi Tay},
  title        = {Sharpness-Aware Minimization Improves Language Model Generalization},
  booktitle    = {{ACL} {(1)}},
  pages        = {7360--7371},
  publisher    = {Association for Computational Linguistics},
  year         = {2022}
}

@article{ntkPINN,
  title={When and why PINNs fail to train: A neural tangent kernel perspective},
  author={Wang, Sifan and Yu, Xinling and Perdikaris, Paris},
  journal={Journal of Computational Physics},
  volume={449},
  pages={110768},
  year={2022},
  publisher={Elsevier}
}

@article{roPINN,
  title={Ropinn: Region optimized physics-informed neural networks},
  author={Wu, Haixu and Luo, Huakun and Ma, Yuezhou and Wang, Jianmin and Long, Mingsheng},
  journal={Advances in Neural Information Processing Systems},
  volume={37},
  pages={110494--110532},
  year={2024}
}

@article{azizzadenesheli2024neural,
  title={Neural operators for accelerating scientific simulations and design},
  author={Azizzadenesheli, Kamyar and Kovachki, Nikola and Li, Zongyi and Liu-Schiaffini, Miguel and Kossaifi, Jean and Anandkumar, Anima},
  journal={Nature Reviews Physics},
  volume={6},
  number={5},
  pages={320--328},
  year={2024},
  publisher={Nature Publishing Group UK London}
}

@inproceedings{saxe2014exact,
  title={Exact solutions to the nonlinear dynamics of learning in deep linear neural networks},
  author={Saxe, A and McClelland, J and Ganguli, S},
  booktitle={Proceedings of the International Conference on Learning Represenatations 2014},
  year={2014},
  organization={International Conference on Learning Represenatations 2014}
}

@inproceedings{damian2022self,
  author       = {Alex Damian and
                  Eshaan Nichani and
                  Jason D. Lee},
  title        = {Self-Stabilization: The Implicit Bias of Gradient Descent at the Edge
                  of Stability},
  booktitle    = {{ICLR}},
  publisher    = {OpenReview.net},
  year         = {2023}
}

@article{karniadakis2021physics,
  title={Physics-informed machine learning},
  author={Karniadakis, George Em and Kevrekidis, Ioannis G and Lu, Lu and Perdikaris, Paris and Wang, Sifan and Yang, Liu},
  journal={Nature Reviews Physics},
  volume={3},
  number={6},
  pages={422--440},
  year={2021},
  publisher={Nature Publishing Group UK London}
}

@article{cheng2024pinnsqp,
  title={Physics-informed neural networks with trust-region sequential quadratic programming},
  author={Cheng, Xiaoran and Na, Sen},
  journal={arXiv preprint arXiv:2409.10777},
  year={2024}
}

@InProceedings{baity2018aging_reference,
  title = 	 {Comparing Dynamics: Deep Neural Networks versus Glassy Systems},
  author =       {Baity-Jesi, Marco and Sagun, Levent and Geiger, Mario and Spigler, Stefano and Arous, Gerard Ben and Cammarota, Chiara and LeCun, Yann and Wyart, Matthieu and Biroli, Giulio},
  booktitle = 	 {Proceedings of the 35th International Conference on Machine Learning},
  pages = 	 {314--323},
  year = 	 {2018},
  volume = 	 {80},
  series = 	 {Proceedings of Machine Learning Research},
  month = 	 {10--15 Jul},
  publisher =    {PMLR},
}

@article{zhu2019physics,
  title={Physics-constrained deep learning for high-dimensional surrogate modeling and uncertainty quantification without labeled data},
  author={Zhu, Yinhao and Zabaras, Nicholas and Koutsourelakis, Phaedon-Stelios and Perdikaris, Paris},
  journal={Journal of Computational Physics},
  volume={394},
  pages={56--81},
  year={2019},
  publisher={Elsevier}
}

@article{ren2022phycrnet,
  title={PhyCRNet: Physics-informed convolutional-recurrent network for solving spatiotemporal PDEs},
  author={Ren, Pu and Rao, Chengping and Liu, Yang and Wang, Jian-Xun and Sun, Hao},
  journal={Computer Methods in Applied Mechanics and Engineering},
  volume={389},
  pages={114399},
  year={2022},
  publisher={Elsevier}
}

@article{gao2021phygeonet,
  title={PhyGeoNet: Physics-informed geometry-adaptive convolutional neural networks for solving parameterized steady-state PDEs on irregular domain},
  author={Gao, Han and Sun, Luning and Wang, Jian-Xun},
  journal={Journal of Computational Physics},
  volume={428},
  pages={110079},
  year={2021},
  publisher={Elsevier}
}

@article{raonic2023convolutional,
  title={Convolutional neural operators for robust and accurate learning of PDEs},
  author={Raonic, Bogdan and Molinaro, Roberto and De Ryck, Tim and Rohner, Tobias and Bartolucci, Francesca and Alaifari, Rima and Mishra, Siddhartha and de B{\'e}zenac, Emmanuel},
  journal={Advances in Neural Information Processing Systems},
  volume={36},
  pages={77187--77200},
  year={2023}
}

@article{li2024physics,
  title={Physics-informed neural operator for learning partial differential equations},
  author={Li, Zongyi and Zheng, Hongkai and Kovachki, Nikola and Jin, David and Chen, Haoxuan and Liu, Burigede and Azizzadenesheli, Kamyar and Anandkumar, Anima},
  journal={ACM/IMS Journal of Data Science},
  volume={1},
  number={3},
  pages={1--27},
  year={2024},
  publisher={ACM New York, NY}
}

@book{engel2001statistical,
  title={Statistical mechanics of learning},
  author={Engel, Andreas},
  year={2001},
  publisher={Cambridge University Press}
}

@article{zdeborova2016statistical,
  title={Statistical physics of inference: Thresholds and algorithms},
  author={Zdeborov{\'a}, Lenka and Krzakala, Florent},
  journal={Advances in Physics},
  volume={65},
  number={5},
  pages={453--552},
  year={2016},
  publisher={Taylor \& Francis}
}

@article{gong2025makes,
  title={What Makes Looped Transformers Perform Better Than Non-Recursive Ones (Provably)},
  author={Gong, Zixuan and Teng, Jiaye and Liu, Yong},
  journal={arXiv preprint arXiv:2510.10089},
  year={2025}
}

@article{lu2024sharpness,
  title={Sharpness-diversity tradeoff: improving flat ensembles with SharpBalance},
  author={Lu, Haiquan and Liu, Xiaotian and Zhou, Yefan and Li, Qunli and Keutzer, Kurt and Mahoney, Michael W and Yan, Yujun and Yang, Huanrui and Yang, Yaoqing},
  journal={Advances in Neural Information Processing Systems},
  volume={37},
  pages={136727--136762},
  year={2024}
}

@inproceedings{hodgkinson2022generalization,
  title={Generalization bounds using lower tail exponents in stochastic optimizers},
  author={Hodgkinson, Liam and Simsekli, Umut and Khanna, Rajiv and Mahoney, Michael},
  booktitle={International Conference on Machine Learning},
  pages={8774--8795},
  year={2022},
  organization={PMLR}
}

@inproceedings{cohen2021gradient,
  author       = {Jeremy Cohen and
                  Simran Kaur and
                  Yuanzhi Li and
                  J. Zico Kolter and
                  Ameet Talwalkar},
  title        = {Gradient Descent on Neural Networks Typically Occurs at the Edge of
                  Stability},
  booktitle    = {{ICLR}},
  publisher    = {OpenReview.net},
  year         = {2021}
}

@article{sagun2017empirical,
  title={Empirical analysis of the hessian of over-parametrized neural networks},
  author={Sagun, Levent and Evci, Utku and Guney, V Ugur and Dauphin, Yann and Bottou, Leon},
  journal={arXiv preprint arXiv:1706.04454},
  year={2017}
}

@article{golatkar2019time,
  title={Time matters in regularizing deep networks: Weight decay and data augmentation affect early learning dynamics, matter little near convergence},
  author={Golatkar, Aditya Sharad and Achille, Alessandro and Soatto, Stefano},
  journal={Advances in Neural Information Processing Systems},
  volume={32},
  year={2019}
}

@article{zhou2023temperature,
  title={Temperature balancing, layer-wise weight analysis, and neural network training},
  author={Zhou, Yefan and Pang, Tianyu and Liu, Keqin and Mahoney, Michael W and Yang, Yaoqing and others},
  journal={Advances in Neural Information Processing Systems},
  volume={36},
  pages={63542--63572},
  year={2023}
}

@inproceedings{negiar2022learning,
  author       = {Geoffrey N{\'{e}}giar and
                  Michael W. Mahoney and
                  Aditi S. Krishnapriyan},
  title        = {Learning differentiable solvers for systems with hard constraints},
  booktitle    = {{ICLR}},
  publisher    = {OpenReview.net},
  year         = {2023}
}

@inproceedings{hansen2023learning,
  title={Learning physical models that can respect conservation laws},
  author={Hansen, Derek and Maddix, Danielle C and Alizadeh, Shima and Gupta, Gaurav and Mahoney, Michael W},
  booktitle={International Conference on Machine Learning},
  pages={12469--12510},
  year={2023},
  organization={PMLR}
}

@article{sakarvadia2025false,
  title={The false promise of zero-shot super-resolution in machine-learned operators},
  author={Sakarvadia, Mansi and Hegazy, Kareem and Totounferoush, Amin and Chard, Kyle and Yang, Yaoqing and Foster, Ian and Mahoney, Michael W},
  journal={arXiv preprint arXiv:2510.06646},
  year={2025}
}

@article{neurde,
  title={Neural equilibria for long-term prediction of nonlinear conservation laws},
  author={Benitez, J and Hegazy, Kareem and Guo, Junyi and Dokmani{\'c}, Ivan and Mahoney, Michael W and de Hoop, Maarten V},
  journal={arXiv preprint arXiv:2501.06933},
  year={2025}
}

@article{chen2018neural,
  title={Neural ordinary differential equations},
  author={Chen, Ricky TQ and Rubanova, Yulia and Bettencourt, Jesse and Duvenaud, David K},
  journal={Advances in neural information processing systems},
  volume={31},
  year={2018}
}

@article{lu2021learning,
  title={Learning nonlinear operators via DeepONet based on the universal approximation theorem of operators},
  author={Lu, Lu and Jin, Pengzhan and Pang, Guofei and Zhang, Zhongqiang and Karniadakis, George Em},
  journal={Nature machine intelligence},
  volume={3},
  number={3},
  pages={218--229},
  year={2021},
  publisher={Nature Publishing Group UK London}
}

@article{krishnapriyan2023learning,
  title={Learning continuous models for continuous physics},
  author={Krishnapriyan, Aditi S and Queiruga, Alejandro F and Erichson, N Benjamin and Mahoney, Michael W},
  journal={Communications Physics},
  volume={6},
  number={1},
  pages={319},
  year={2023},
  publisher={Nature Publishing Group UK London}
}

@inproceedings{liu2024model,
  author       = {Zihang Liu and
                  Yuanzhe Hu and
                  Tianyu Pang and
                  Yefan Zhou and
                  Pu Ren and
                  Yaoqing Yang},
  title        = {Model Balancing Helps Low-data Training and Fine-tuning},
  booktitle    = {{EMNLP}},
  pages        = {1311--1331},
  publisher    = {Association for Computational Linguistics},
  year         = {2024}
}

@article{kidger2020neural,
  title={Neural controlled differential equations for irregular time series},
  author={Kidger, Patrick and Morrill, James and Foster, James and Lyons, Terry},
  journal={Advances in neural information processing systems},
  volume={33},
  pages={6696--6707},
  year={2020}
}

@article{batzner20223,
  title={E (3)-equivariant graph neural networks for data-efficient and accurate interatomic potentials},
  author={Batzner, Simon and Musaelian, Albert and Sun, Lixin and Geiger, Mario and Mailoa, Jonathan P and Kornbluth, Mordechai and Molinari, Nicola and Smidt, Tess E and Kozinsky, Boris},
  journal={Nature communications},
  volume={13},
  number={1},
  pages={2453},
  year={2022},
  publisher={Nature Publishing Group UK London}
}

@inproceedings{satorras2021n,
  title={E (n) equivariant graph neural networks},
  author={Satorras, V{\i}ctor Garcia and Hoogeboom, Emiel and Welling, Max},
  booktitle={International conference on machine learning},
  pages={9323--9332},
  year={2021},
  organization={PMLR}
}

@article{rao2023encoding,
  title={Encoding physics to learn reaction--diffusion processes},
  author={Rao, Chengping and Ren, Pu and Wang, Qi and Buyukozturk, Oral and Sun, Hao and Liu, Yang},
  journal={Nature Machine Intelligence},
  volume={5},
  number={7},
  pages={765--779},
  year={2023},
  publisher={Nature Publishing Group UK London}
}

@article{wang2023scientific,
  title={Scientific discovery in the age of artificial intelligence},
  author={Wang, Hanchen and Fu, Tianfan and Du, Yuanqi and Gao, Wenhao and Huang, Kexin and Liu, Ziming and Chandak, Payal and Liu, Shengchao and Van Katwyk, Peter and Deac, Andreea and others},
  journal={Nature},
  volume={620},
  number={7972},
  pages={47--60},
  year={2023},
  publisher={Nature Publishing Group UK London}
}

@inproceedings{adam2014method,
  author       = {Diederik P. Kingma and
                  Jimmy Ba},
  title        = {Adam: {A} Method for Stochastic Optimization},
  booktitle    = {{ICLR}},
  year         = {2015}
}

@inproceedings{hu2025eigenspectrum,
  title = 	 {Eigenspectrum Analysis of Neural Networks without Aspect Ratio Bias},
  author =       {Hu, Yuanzhe and Goel, Kinshuk and Killiakov, Vlad and Yang, Yaoqing},
  booktitle = 	 {Proceedings of the 42nd International Conference on Machine Learning},
  pages = 	 {24290--24313},
  year = 	 {2025},
  volume = 	 {267},
  series = 	 {Proceedings of Machine Learning Research},
  month = 	 {13--19 Jul},
  publisher =    {PMLR},
}

@inproceedings{he2015deepresiduallearningimage,
  author       = {Kaiming He and
                  Xiangyu Zhang and
                  Shaoqing Ren and
                  Jian Sun},
  title        = {Deep Residual Learning for Image Recognition},
  booktitle    = {{CVPR}},
  pages        = {770--778},
  publisher    = {{IEEE} Computer Society},
  year         = {2016}
}

@inproceedings{dosovitskiy2021imageworth16x16words,
  author       = {Alexey Dosovitskiy and
                  Lucas Beyer and
                  Alexander Kolesnikov and
                  Dirk Weissenborn and
                  Xiaohua Zhai and
                  Thomas Unterthiner and
                  Mostafa Dehghani and
                  Matthias Minderer and
                  Georg Heigold and
                  Sylvain Gelly and
                  Jakob Uszkoreit and
                  Neil Houlsby},
  title        = {An Image is Worth 16x16 Words: Transformers for Image Recognition
                  at Scale},
  booktitle    = {{ICLR}},
  publisher    = {OpenReview.net},
  year         = {2021}
}

@inproceedings{woo2023convnextv2codesigningscaling,
  author       = {Sanghyun Woo and
                  Shoubhik Debnath and
                  Ronghang Hu and
                  Xinlei Chen and
                  Zhuang Liu and
                  In So Kweon and
                  Saining Xie},
  title        = {ConvNeXt {V2:} Co-designing and Scaling ConvNets with Masked Autoencoders},
  booktitle    = {{CVPR}},
  pages        = {16133--16142},
  publisher    = {{IEEE}},
  year         = {2023}
}

@article{zhu1997algorithm,
  title={Algorithm 778: L-BFGS-B: Fortran subroutines for large-scale bound-constrained optimization},
  author={Zhu, Ciyou and Byrd, Richard H and Lu, Peihuang and Nocedal, Jorge},
  journal={ACM Transactions on mathematical software (TOMS)},
  volume={23},
  number={4},
  pages={550--560},
  year={1997},
  publisher={ACM New York, NY, USA}
}

@inproceedings{galvis2017overlapping,
  title={On overlapping domain decomposition methods for high-contrast multiscale problems},
  author={Galvis, Juan and Chung, Eric T and Efendiev, Yalchin and Leung, Wing Tat},
  booktitle={International Conference on Domain Decomposition Methods},
  pages={45--57},
  year={2017},
  organization={Springer}
}

@Article{energy_landscape_wales17,
   author  = "A. J. Ballard and R. Das and S. Martiniani and D. Mehta and L. Sagun and J. D. Stevenson and D. J. Wales",
   title   = "Energy landscapes for machine learning",
   journal = "Physical Chemistry Chemical Physics",
   year    = "2017",
   volume  = "19",
   issue   = "20",
   pages   = "12585-12603",
}

@TECHREPORT{MM17_TR,
  author =    {C. H. Martin and M. W. Mahoney},
  title =     {Rethinking generalization requires revisiting old ideas: statistical mechanics approaches and complex learning behavior},
  number =    {Preprint: arXiv:1710.09553},
  year =      {2017},
}

@article{bartlett2020benign,
  title={Benign overfitting in linear regression},
  author={Bartlett, Peter L and Long, Philip M and Lugosi, G{\'a}bor and Tsigler, Alexander},
  journal={Proceedings of the National Academy of Sciences},
  volume={117},
  number={48},
  pages={30063--30070},
  year={2020},
  publisher={National Academy of Sciences}
}

@article{hastie2022surprises,
  title={Surprises in high-dimensional ridgeless least squares interpolation},
  author={Hastie, Trevor and Montanari, Andrea and Rosset, Saharon and Tibshirani, Ryan J},
  journal={Annals of statistics},
  volume={50},
  number={2},
  pages={949},
  year={2022}
}

@article{belkin2019reconciling,
  title={Reconciling modern machine-learning practice and the classical bias--variance trade-off},
  author={Belkin, Mikhail and Hsu, Daniel and Ma, Siyuan and Mandal, Soumik},
  journal={Proceedings of the National Academy of Sciences},
  volume={116},
  number={32},
  pages={15849--15854},
  year={2019},
  publisher={National Academy of Sciences}
}
\bibliographystyle{./icml-2026/icml2026}

\clearpage
\appendix
\onecolumn
\section{Related Work}
\label{app:related_work}

\paragraph{Machine Learning Differential Equations.} 
ML-based approaches for ``solving'' differential equations have received attention in recent years~\citep{karniadakis2021physics,azizzadenesheli2024neural}. 
One popular approach is based on PINNs~\citep{raissi2019unified,krishnapriyan2021characterizing} and its variants~\citep{zhu2019physics,gao2021phygeonet,ren2022phycrnet}, which exploit soft Lagrangian relaxation as a penalty method to incorporate PDE residuals as soft constraints. 
Since such formulations can be ill-conditioned and very sensitive to optimization hyperparameters, a range of optimization~\citep{ntkPINN,roPINN,nncg} and hard-constraint methods~\citep{lu2021physics,cheng2024pinnsqp} have been proposed to help improve training stability. 
Relatedly, to enable instance-level generalization, NOs~\citep{li2020fourier,lu2021learning,raonic2023convolutional,sakarvadia2025false} aim to learn mappings between functional spaces for instance-level generalization. 
In a complementary direction, NODEs~\citep{chen2018neural,kidger2020neural,krishnapriyan2023learning} are motivated by continuous-time modeling, by learning the time derivative with a neural network (NN) and integrating it with a differentiable ODE solver. 
Despite their popularity, the training and generalization properties of these models can be extremely sensitive to design and optimization choices~\citep{krishnapriyan2021characterizing,krishnapriyan2023learning,sakarvadia2025false}. 
This has motivated structure-preserving approaches that encode physical structure as stronger inductive biases, often by incorporating numerical principles at macro-scale~\citep{negiar2022learning,hansen2023learning,rao2023encoding}, meso-scale~\citep{neurde}, or micro-scale~\citep{satorras2021n,batzner20223} levels.

\paragraph{Loss Landscape Analysis.}  
Loss landscape analysis comes from chemical physics~\citep{energy_landscape_wales17}, and it has been widely used to understand various phenomena in NN models. 
Within CV, it has been instrumental in ``explaining'' normalization and residual connections~\citep{li2018visualizing, yao2020pyhessian}, large-batch training~\citep{yao2018large}, sharpness-aware optimization~\citep{foret2020sharpness}, and large-scale characterization of NN behavior~\citep{yang2021taxonomizing}. 
Similarly, within NLP, it has been used to study the stage-wise spectral degeneration of transformers~\citep{hooglandloss} as well as pre-training and fine-tuning~\citep{chen2025understanding}. 
Loss landscape analysis has also been applied in SciML to characterize ill-conditioning properties of PINNs~\citep{krishnapriyan2021characterizing, xie2024losslens, nncg, roPINN}. 
Methods for loss landscape analysis have advanced significantly in recent years. 
Early work mainly focused on visualization~\citep{li2018visualizing}, but subsequent studies introduced quantitative local measures by sharpness and Hessian-based metrics~\citep{yao2020pyhessian}. 
Motivated by going beyond purely local geometry, subsequent work also developed more global characterizations of the landscape, such as mode connectivity~\citep{garipov2018loss,draxler2018essentially}.
These tools have been used to classify phase-transitions~\citep{yang2021taxonomizing} and to inform hyperparameter selection in training models~\citep{zhou2023temperature,liu2024model,hu2025eigenspectrum}. More recently, topological data analysis has been explored as an additional lens for probing loss-landscape structure~\citep{xie2024losslens}.

\paragraph{Training Regimes in Optimization.} 
NN training is known to exhibit distinct \emph{phases}. 
The characterization of these phases is rooted in statistical physics~\citep{engel2001statistical, zdeborova2016statistical,MM17_TR}, where phase transitions often manifest as discontinuous changes in system properties, governed by control parameters such as optimization noise and the data-model ratio. 
In ML, such transitions are often associated with generalization, e.g., ``double descent'' behavior, observed as model size or epoch count increases \citep{bartlett2020benign,nakkiran2019deep,hastie2022surprises,belkin2019reconciling}. 
These macroscopic regimes are intrinsically tied to the geometry of the loss landscape, where the optimizer's trajectory is dictated by the connectivity of local basins~\citep{yang2021taxonomizing}. 
While recent studies debate whether certain transitions in large language models are indeed emergent or induced by specific metrics~\citep{wei2022emergent, schaeffer2024emergent}, the phase-based studies still provide a crucial practical way of diagnosing optimization~failures.

\section{Problem Setup}
\label{app:prob_setup}
In this paper, the focus is on SciML methods for learning differential equation systems. This section outlines the basic methodology of standard SciML models and their optimization problems, including both data-driven approaches (e.g., NODEs and FNOs) and physics-constrained methods (e.g., PINNs, PINOs, and PINODEs).

\subsection{SciML Models}
\label{app:models}

Here, we describe the basic SciML models we used in our analysis.

\paragraph{PINNs.}
The main goal of PINNs is to solve PDEs. Similar to prior work~\citep{hao2023pinnacle}, we consider the general setup of PDEs:
\begin{subequations}
    \label{eq:gen_pde}
    \begin{align}
        & \Dc[u(x),x] = 0, \quad x\in \Omega, \\
        & \Bc[u(x),x] = 0, \quad x\in \partial \Omega, 
    \end{align} 
\end{subequations}
where $\Dc$ is a differential operator defining the PDE, $\Bc$ is an operator associated with the boundary and/or initial conditions, and $\Omega \subseteq \R^d$. PINNs model $u$ as an NN and approximate the true solution by the network whose weights solve the following non-linear least-squares problem:
\begin{align}
\label{eq:pinn_prob_gen}
    \underset{w\in \R^p}{\mbox{minimize}}~L(w) \coloneqq & \frac{1}{2\nres}\sum_{i=1}^{\nres}\left(\Dc[u(x_r^i; w),x_r^i]\right)^2 
    +\frac{1}{2\nbc}\sum^{\nbc}_{j=1}\left(\Bc[u(x_b^j;w),x_b^j]\right)^2. 
\end{align}
Here, $\nres$ and $\nbc$ denote the numbers of residual and boundary/initial points, $\{x_r^i\}^{\nres}_{i=1}$ are the residual points, and $\{x^j_b\}^{\nbc}_{j=1}$ are the boundary/initial points.
The first loss term measures how much $u(x;w)$ fails to satisfy the PDE, while the second term measures how much $u(x;w)$ fails to satisfy the boundary/initial conditions.

\paragraph{NOs and PINOs.}
NOs~\citep{li2020neural,lu2021learning} extend standard NNs by learning mappings between function spaces. Specifically, they approximate an operator $G: \mathcal{A} \rightarrow \mathcal{U}$, where $\mathcal{A}$ denotes an input function space (e.g., initial conditions, boundary conditions, or PDE coefficients) and $\mathcal{U}$ denotes the corresponding solution space. The model parameterizes a mapping $G_\theta$ such that
\begin{equation}
    u = G_\theta(a), \quad a \in \mathcal{A},
\end{equation}
where $\theta$ denotes the trainable parameters. In practice, functions are discretized on a grid, and $a_i \in \mathbb{R}^{d_a}$ and $u_i \in \mathbb{R}^{d_u}$ denote the discretized input and solution fields for the $i$-th sample. The operator is trained by minimizing a data-driven loss,
\begin{equation}
\label{eq:no_loss}
    \mathcal{L}_{\text{data}}(\theta) = \frac{1}{N} \sum_{i=1}^{N} \| G_\theta(a_i) - u_i \|^2,
\end{equation}
where $\{(a_i, u_i)\}_{i=1}^N$ are input–output pairs, and $\|\cdot\|$ denotes the Euclidean norm over the discretized field.

For physics-informed variants (PINOs), additional constraints are incorporated to enforce consistency with governing equations. This is achieved by augmenting the objective with a residual loss defined on collocation points,
\begin{equation}
    \mathcal{L}_{\text{phys}}(\theta) = \frac{1}{M} \sum_{j=1}^{M} \| \mathcal{F}(G_\theta(a)(x_j)) \|^2,
\end{equation}
where $\mathcal{F}$ denotes the differential operator associated with the governing PDE and $\{x_j\}_{j=1}^M$ are collocation points in the spatial (and possibly temporal) domain. The final loss function is given by,
\begin{equation}
    \mathcal{L}(\theta) = \mathcal{L}_{\text{data}}(\theta) + \lambda \mathcal{L}_{\text{phys}}(\theta),
\end{equation}
with $\lambda$ is the weighting coefficient.

\paragraph{NODEs and PINODEs.}
We consider NODEs~\citep{chen2018neural} for modeling temporal dynamics of the form
\begin{equation}
    \frac{dx}{dt} = f_\theta(x, t),
\end{equation}
where $f_\theta$ is parameterized by NNs. Given an initial condition $x(0)$, predictions at time $t_i$ are obtained via numerical integration,
\begin{equation}
    \hat{x}(t_i) = x(0) + \int_{0}^{t_i} f_\theta(x(\tau), \tau)\, d\tau,
\end{equation}
which is approximated using a fixed-step solver (e.g., forward Euler) aligned with the temporal resolution of the training data. The standard NODE is trained by minimizing the mean squared error (MSE) between predicted and observed trajectories,
\begin{equation}
    \mathcal{L}_{\text{data}} = \frac{1}{N} \sum_{i=1}^N \|\hat{x}(t_i; \theta) - x(t_i)\|^2,
\end{equation}
where $\{t_i\}_{i=1}^N$ denotes the set of observation time points, $x(t_i) \in \mathbb{R}^d$ is the ground-truth state at time $t_i$, and $\hat{x}(t_i; \theta)$ is the model prediction obtained by integrating the learned dynamics $f_\theta$. The norm $\|\cdot\|$ denotes the Euclidean norm in $\mathbb{R}^d$.

For the physics-informed variant, additional loss terms are incorporated by enforcing consistency with known governing equations or physical constraints. This is achieved by augmenting the objective with a residual loss,
\begin{equation}
    \mathcal{L}_{\text{phys}} = \frac{1}{M} \sum_{j=1}^M \| \partial_t \hat{x}(t_j) - \mathcal{F}(\hat{x}(t_j)) \|^2,
\end{equation}
where $\mathcal{F}$ represents the known dynamics (when available). The overall objective combines data fidelity and physical consistency,
\begin{equation}
    \mathcal{L} = \mathcal{L}_{\text{data}} + \lambda \mathcal{L}_{\text{phys}},
\end{equation}
with $\lambda$ controlling the trade-off between the two terms.

\subsection{Differential Equations}
\label{app:diff_equations}
Here, we describe the differential equations considered in the experimental studies. We use: (1) 1D convection, reaction, wave, and reaction-diffusion equations for PINNs; (2) 2D Poisson and advection-diffusion equations for NOs; (3) 2D Darcy flow for PINOs; and (4) the nonlinear Pendulum for NODEs and PINODEs.

\paragraph{1D Convection.}
The 1D convection problem is a hyperbolic PDE that can be used to model fluid flow, heat transfer, and biological processes. We consider the setup of the convection equation in~\citet{krishnapriyan2021characterizing}, given by 
\begin{align*}
    \frac{\partial u}{\partial t} + \beta \frac{\partial u}{\partial x} = 0, & \quad x \in [0, 2\pi), t \in (0, 1), \\
    u(x, 0) = \sin(x), & \quad u(0, t) = u(2 \pi, t). 
\end{align*}
The analytical solution to this PDE is $u(x, t) = \sin(x - \beta t)$. 
In this paper, the parameter $\beta$ is varied over the range $[5,70]$.

\paragraph{1D Reaction.}
The 1D reaction problem is governed by a nonlinear PDE commonly used to model chemical reaction dynamics. The specific reaction PDE considered in this study is given by~\citep{roPINN}:
\begin{align*}
    \frac{\partial u}{\partial t} - \rho u (1 - u) = 0, & \quad x \in (0, 2\pi), t \in (0, 1) \\
    u(x, 0) = \exp \left( -\frac{(x - \pi)^2}{2 (\pi / 4)^2} \right), & \quad x \in [0, 2\pi], \\
    u(0, t) = u(2 \pi, t), & \quad t \in [0, 1].
\end{align*}
The analytical solution to this ODE is $u(x, t) = \frac{h(x) e^{\rho t}}{h(x) e^{\rho t} + 1 - h(x)}$, where $h(x) = \exp \left( -\frac{(x - \pi)^2}{2 (\pi / 4)^2} \right)$.

\paragraph{1D Wave.}
The 1D wave equation is widely used to model wave propagation phenomena in acoustics, elasticity, and electromagnetics. We follow the setup in~\citet{roPINN}:
\begin{align*}
    \label{eq:1dwave}
    \frac{\partial^2 u}{\partial t^2} - c^2 \cdot \frac{\partial^2 u}{\partial x^2} = 0, & \quad x \in (0,1), t \in(0,1), \\
    u(x,0) = \text{sin}(\pi x) + 0.5 \text{sin}(\beta \pi x), & \quad x \in [0,1], \\
    \frac{\partial u(x,0)}{\partial t} = 0, & \quad x \in [0,1], \\
    u(0,t) = u(1,t) = 0, & \quad t \in [0,1].
\end{align*}

\paragraph{1D Reaction-Diffusion.}
The 1D reaction-diffusion equation is a prototypical nonlinear PDE that models the interplay between diffusion-driven transport and local reaction kinetics. Following the setup in~\citet{krishnapriyan2021characterizing}, we formulate this equation as:
\begin{equation*}
    \label{eq:1drd}
    \frac{\partial u}{\partial t} = \nu \frac{\partial^2 u}{\partial x^2} - \rho u(1-u), \quad x \in [0, 2\pi), t \in (0,1).
\end{equation*}

\paragraph{2D Poisson.} 
Here, we adopt the same PDE as SYS-1 in ~\citet{subramanian2023towards}. Specifically, we consider an elliptic system with periodic boundary conditions on the domain $\Omega = [0, 1]^2$: 
\begin{equation}
    -\operatorname{div} \mathbf{K} \nabla u = f, \quad \text{in}~\Omega,
\end{equation}
where $u$ denotes the state variable, $f(\mathbf{x})$ is a source (forcing) term, and $\mathbf{K}$ is the diffusion coefficient tensor. The tensor $\mathbf{K}$ encodes the underlying material or transport physics of the system. The parameter $\mathbf{K}$ is selected within $[1,200]$ throughout this study.

\paragraph{2D Advection-Diffusion.}
Following the SYS-2 setup in~\citet{subramanian2023towards}, we further evaluate a more complex 2D advection-diffusion PDE, given by  
\begin{equation}
    -\operatorname{div} \mathbf{K} \nabla \mathbf{u} + \boldsymbol{\nu} \cdot \nabla \mathbf{u} = f, \quad \text{in}~\Omega,
\end{equation}
where $\boldsymbol{\nu}$ denotes the velocity vector. We use the ratio of advection to diffusion as a varying coefficient: $\boldsymbol{\Psi} = \| \boldsymbol{\nu} \cdot \nabla \mathbf{u} \| / \| \operatorname{div} \mathbf{K} \nabla \mathbf{u} \|$.

\paragraph{2D Darcy Flow.}
We consider the steady-state Darcy flow equation~\citep{li2024physics} on the unit square domain $\Omega=(0,1)^2$ with the homogeneous Dirichlet boundary condition. The governing PDE is
\begin{align}
-\nabla \cdot \big(a(x)\nabla u(x)\big) &= f(x), \quad x \in \Omega, \\
u(x) &= 0, \quad x \in \partial \Omega,
\end{align}
where $a(x)\in L^\infty(\Omega;\mathbb{R}_+)$ denotes a piecewise-constant diffusion coefficient and $f(x)=1$ is a fixed forcing term. The task is to learn the nonlinear operator that maps the diffusion field $a$ to the corresponding solution field $u$. For the physics-informed training objective, we use the PDE residual
\begin{equation}
\mathcal{L}_{\text{pde}}(u)
=
\nabla \cdot \big(a\nabla u\big)-f.
\end{equation}

To control the physical difficulty of the problem, we vary the contrast ratio~\citep{galvis2017overlapping},
\begin{equation}
r = \frac{a_{\max}}{a_{\min}},
\end{equation}
which measures the scale separation between the largest and smallest diffusion coefficients. Larger contrast ratios correspond to stronger multiscale heterogeneity and higher physical stiffness, making the operator-learning problem more challenging.

\paragraph{Nonlinear Damped Pendulum.}
We study a nonlinear damped pendulum to probe training dynamics of time-continuous models under varying physical complexity and data availability. The system is governed by
\begin{equation}
    \label{eq:pendulum_dynamics}
    \frac{d^2\theta}{dt^2} + b \frac{d\theta}{dt} + \sin(\theta) = 0,
\end{equation}
or equivalently, the first-order form
\begin{equation}
    \dot{\theta} = \omega,\qquad \dot{\omega} = -b\,\omega - \sin(\theta),
\end{equation}
where $\theta(t)$ is the angle and $b \in [0.1, 1.0]$ is the damping coefficient. Smaller $b$ yields longer-lived, more oscillatory nonlinear dynamics, while larger $b$ produces rapidly decaying trajectories.

\section{Extended Experiment Setups}
\label{app:extended_experiment_setup}

\subsection{Summary of Experimental Configurations}
\label{app:summary_experiment_config}

Table~\ref{tab:experiment_setup_summary} summarizes the representative SciML models, benchmark physical systems, and optimization or training strategies evaluated in this work. The selected models span several major SciML paradigms, including collocation-based physics-constrained learning (PINNs), operator learning (FNOs), physics-informed operator learning (PINOs), and temporal dynamics models (NODEs and PINODEs). The benchmark systems cover representative PDE and ODE settings with varying physical difficulty, including transport-dominated dynamics, elliptic systems, multiscale porous-media flow, and nonlinear dynamical systems. We evaluate a broad set of optimization and training interventions, including standard first- and second-order optimization methods, hard-constrained formulations, curriculum learning strategies, and PINN-specific stabilization methods. These experiments form the basis of our regime-aware empirical analysis.

\begin{table*}[ht!]
\centering
\caption{Summary of SciML models, benchmark systems, and optimization/training strategies evaluated in this work.}
\label{tab:experiment_setup_summary}
\resizebox{0.85\textwidth}{!}{
\begin{tabular}{lll}
\toprule
\textbf{SciML Model} & \textbf{Representative Physical System} & \textbf{Optimization / Training Strategies} \\
\midrule

PINN 
& 1D convection, reaction, wave, reaction-diffusion equations 
& L-BFGS, ALM, NNCG, RoPINN, CL \\

FNO 
& 2D Poisson, advection-diffusion equations 
& Adam \\

PINO 
& 2D Darcy flow 
& Adam, L-BFGS, ALM, NNCG, CL \\

NODE 
& Nonlinear pendulum system 
& Adam \\

PINODE 
& Nonlinear pendulum system 
& Adam, L-BFGS, ALM, NNCG, CL \\

\bottomrule
\end{tabular}
}
\end{table*}

\subsection{Training Protocols}
\label{app:training_protocols}

\paragraph{PINNs.}
For the PINN experiments, we sweep the governing physical parameter of each PDE system to control the underlying physical difficulty and construct regime plots. For the 1D convection equation, the convection coefficient is varied over $\beta \in [5,70]$. For the reaction equation, we vary the reaction coefficient over $\rho \in [1,12]$. For the wave equation, the wave speed is swept over $c \in [0.1,6]$. Finally, for the reaction-diffusion equation, we evaluate regimes across $\rho \in [1,30]$. In all cases, increasing the physical parameter generally produces sharper solution structures, stronger stiffness, or more challenging multiscale behavior, thereby increasing the optimization and generalization difficulty of the corresponding SciML problem.
The PINN architecture uses a Multi-Layer Perceptron (MLP) with four hidden layers for all related PDE scenarios. The hidden dimensions are $[50, 50, 50, 50]$. We run PINN experiments using five random seeds. The training configurations are provided in Table~\ref{tab:pinn_opt_protocols}.

\begin{table*}[t]
\centering
\caption{Training protocols and optimization configurations for PINN experiments.}
\label{tab:pinn_opt_protocols}
\resizebox{0.92\textwidth}{!}{
\begin{tabular}{lllll}
\toprule
\textbf{Strategy} 
& \textbf{Learning Rate} 
& \textbf{Epochs} 
& \textbf{Batch Size} 
& \textbf{Additional Settings} \\
\midrule

L-BFGS
& 1.0
& $2 \times 10^{3}$
& Full batch
& $\text{history}=100$; $\text{tolerance\_grad}=10^{-7}$, $\text{tolerance\_change}=10^{-9}$ \\

NNCG
& 0.001 / 1.0
& $5 \times 10^{3}$
& Full batch
& Adam: $2 \times 10^{3}$ epochs, L-BFGS: $1 \times 10^{3}$ epochs, NNCG: $2 \times 10^{3}$ epochs; rank = 50 \\

ALM
& 1.0
& $50 \times 2 \times 10^{3}$
& Full batch
& penalty coefficient increases from 2 to $\mu$ = $1.2^{50}$; multiplier update every $2 \times 10^{3}$ epochs \\

RoPINN
& 1.0
& $1 \times 10^{3}$
& Full batch
& init. region = $10^{-5}$; samples = 1; window = 5 \\

CL
& 1.0
& $140 \times 2 \times 10^{3}$
& Full batch
& progressively increases PDE coefficient at speed of 0.05 every $2 \times 10^{3}$ iterations \\

\bottomrule
\end{tabular}
}
\end{table*}

\paragraph{FNOs.} 
For the experimental setup of FNOs, we follow the methodology described in~\citet{subramanian2023towards}. We focus on solving the 2D Poisson (SYS-1) and advection–diffusion (SYS-2) systems under varying parameter configurations. Specifically, we generate multiple sets of SYS-1 by varying the governing parameters across the range: ${(1.0, 2.5), (2.5, 5.0), (5.0, 10), (10, 20), (20, 30), (30, 50), (50, 100), (100, 200)}$ and  SYS-2 with the settings across the range: ${(0.2, 0.4), (0.4, 0.6), (0.6, 0.8), (0.8, 10), (1.0, 1.3), (1.3, 1.7), (1.7, 2.0), (2.0, 2.5)}$.
The FNO model used in our experiments contains approximately one million parameters. Other hyperparameters, including batch size and learning rate, are aligned with those reported in~\citet{subramanian2023towards} to ensure consistency and comparability. To balance convergence and overfitting, we adjust the number of training epochs based on the dataset size. Our settings are as follows, where the first value denotes dataset size (number of PDE samples) and the second value denotes the corresponding number of training epochs: $\{16\text{k}: 75, 8\text{k}: 100, 4\text{k}: 150, 2\text{k}: 200, 1\text{k}: 300, 512: 500, 256: 750, 128: 1000\}$. We use the Adam optimizer for training FNOs across five random seeds.

\paragraph{PINOs.} We follow the procedure and the experimental setup of 2D Darcy Flow in~\citet{li2024physics}. The PINO backbone is an FNO model with spectral modes 20 per spatial dimension, channel width 64, Fourier block layout $[64,64,64,64,64]$. The coefficient field $a(x)$ is generated from a Gaussian Random Field-based pipeline, and we sweep the contrast ratio (physical parameter) in $\{1,2,3,4,5,6,8,10\}$, training set size in $\{5,10,20,50,100,200,500,1000\}$, and use three random seeds. The training details are provided in Table~\ref{tab:pino_opt_protocols}. 

\begin{table*}[t]
\centering
\caption{Training protocols and optimization configurations for PINO experiments.}
\label{tab:pino_opt_protocols}
\resizebox{0.9\textwidth}{!}{
\begin{tabular}{llllll}
\toprule
\textbf{Strategy} 
& \textbf{Learning Rate} 
& \textbf{Epochs} 
& \textbf{Batch Size} 
& \textbf{Additional Settings} \\
\midrule

Adam
& 0.001
& $1 \times 10^{4}$
& 32
& $(\beta_1,\beta_2)=(0.9,0.999)$ \\

L-BFGS
& 1.0
& $1 \times 10^{4}$
& Full batch
& $\text{history}=100$; $\text{tolerance\_grad}=10^{-7}$, $\text{tolerance\_change}=10^{-9}$ \\

NNCG
& 0.001 / 1.0
& $1.11 \times 10^{4}$
& 32 / Full batch
& Adam: $1 \times 10^{4}$ epochs, L-BFGS: $1 \times 10^{3}$ epochs, NNCG: 100 epochs; rank = 4 \\

ALM
& 0.0001
& $3.5 \times 10^{4}$
& 32
& penalty coefficient increase from 2 to $\mu$ = $1.05^{40}$; multiplier update every $5 \times 10^{2}$ iterations \\

CL
& 0.001
& $100 \times 10^{4}$
& 32
& progressively increases PDE coefficient at speed of 0.1 every $1 \times 10^{4}$ iterations \\

\bottomrule
\end{tabular}
}
\end{table*}

\paragraph{NODEs and PINODEs.}
The network and data setup are the same for both NODEs and PINODEs. 
We use a shallow MLP with three hidden layers (each with a width of 16). Predictions are produced by numerical integration:
\begin{equation}
    \hat{x}(t_i) = x(0) + \int_{0}^{t_i} f_\theta(x(\tau)) \, d\tau.
\end{equation}
During training, we integrate with the forward Euler method using the same step size as the training data ($\Delta t = 0.05$). The model minimizes mean-squared error (MSE) between predicted and ground-truth trajectories over the training horizon. Ground-truth trajectories are generated with \texttt{solve\_ivp} using default tolerances. We embed the phase-space state $[\theta,\omega]$ into $\mathbb{R}^3$ via a spherical map:
\begin{equation}
    x(t) = \begin{bmatrix} 
    \sin(\theta) \cos(\omega) \\ 
    \sin(\theta) \sin(\omega) \\ 
    -\cos(\theta)
    \end{bmatrix},
\end{equation}
which constrains dynamics to the unit sphere and introduces geometric complexity. Training observations are sampled at a fixed temporal resolution $\Delta t = 0.05$. We sweep two axes: (1) System Nonlinearity ($x$-axis): We vary damping $b \in [0.1, 1.0]$ and visualize the $x$-axis as $1/b$ (rightward indicates lower damping and more oscillatory dynamics); (2) Data Availability ($y$-axis): We vary training horizon $T_{\text{train}} \in [1.0, 20.0]$, with number of samples $N_t = T_{\text{train}} / \Delta t$. While $T_{\text{train}}$ varies, evaluation is done on a fixed long horizon $T_{\text{test}} = 20.0$s from a distinct initial condition ($\theta_0^{\text{test}}=2.8$ v.s.\ $\theta_0^{\text{train}}=1.7$). The reported test error is the one-step prediction error along this full test trajectory, probing generalization across initial conditions. We also use Adam for training NODEs. 
Table~\ref{tab:pinode_opt_protocols} also presents the training details of PINODEs for different training and optimization methods.

\begin{table*}[t]
\centering
\caption{Training protocols and optimization configurations for PINODE experiments.}
\label{tab:pinode_opt_protocols}
\resizebox{0.9\textwidth}{!}{
\begin{tabular}{llllll}
\toprule
\textbf{Strategy} 
& \textbf{Learning Rate} 
& \textbf{Epochs} 
& \textbf{Batch Size} 
& \textbf{Additional Settings} \\
\midrule

Adam
& $10^{-3}$
& $6 \times 10^{2}$
& 32
& $(\beta_1,\beta_2)=(0.9,0.999)$ \\

L-BFGS
& 1.0
& $1 \times 10^{3}$
& Full batch
& $\text{history}=50$; $\text{tolerance\_grad}=10^{-7}$, $\text{tolerance\_change}=10^{-9}$ \\

NNCG
& $10^{-3}$
& $1.5 \times 10^{3}$
& Full batch
& LBFGS warmup for $1 \times 10^{3}$ epochs; Nyström rank = 10 \\

ALM
& 1.0
& $50 \times 500$
& Full batch
& penalty coefficient $\mu$ = $1.05^{50}$; multiplier update every $5 \times 10^{2}$ epochs \\

CL
& 1.0
& $60 \times 500$
& Full batch
& progressively increases PDE coefficient at speed of 0.25 every $5 \times 10^{2}$ epochs \\

\bottomrule
\end{tabular}
}
\end{table*}

\section{Implementations of Visualization and Multi-Regime Analysis}
\label{app:implement_loss_landscape_regime}

\subsection{Loss Landscape Metrics}
\label{app:loss_landscape_metrics}

Prior work \citep{yang2021taxonomizing} categorizes phases in load-temperature variations using four metrics. The first metric is the training loss, which evaluates whether the training data is interpolated. The other metrics describe the sharpness of the local minima, the similarity between models trained using different random seeds, and the connectivity between different local minima of the loss landscape. 
It should be noted that~\citet{yang2021taxonomizing} used a certain set of metrics to measure these loss landscape properties, but there are alternative metrics available. For example, the sharpness of local minima can be measured using \emph{adaptive sharpness metrics}~\citep{andriushchenko2023modern,kwon2021asam},
while similarity can be measured using \emph{disagreement}~\citep{theisenWhenAreEnsembles2023}.

We define the loss landscape metrics following~\citet{yang2021taxonomizing}. Let $\vtheta \in \mathbb{R}^m$ denote the learnable weight parameter, and let $\mathcal{L}$ be the loss function. We compute metrics using the train set unless stated otherwise.

\textbf{Hessian-based Metrics.} \
The Hessian matrix $\mathbf{H}$ at a given point $\vtheta$ can be represented as $\nabla^2_\mathbf{\vtheta} \mathcal{L}(\vtheta) \in \mathbb{R}^{m \times m}$. The largest eigenvalue $\lambda_{\max}(\rmH)$ and trace $\text{Tr}(\mathbf{H})$ are used to summarize the local curvature properties in a single value. Specifically, a larger value of the top eigenvalue or trace indicates greater sharpness.

\textbf{Mode Connectivity.} \
The mode connectivity assesses the presence of low-loss paths between different local minima and reflects how well different solutions are connected in the parameter space, indicating smoother and more generalizable loss landscapes. It is common to fit Bézier curves, $(\gamma_{\phi}(t)$, between two models $\vtheta$ and $\vtheta'$, and subsequently compute mode connectivity $\texttt{mc}$ as
\[
\texttt{mc}(\vtheta, \vtheta') = \frac{1}{2}(\mathcal{L}(\vtheta) + \mathcal{L}(\vtheta')) - \mathcal{L}(\gamma_{\phi}(t^*)),
\]
where $t^* = \underset{t}{\mathrm{argmin}} \left| \frac{1}{2}(\mathcal{L}(\vtheta) + \mathcal{L}(\vtheta')) - \mathcal{L}(\gamma_{\phi}(t)) \right|$. 
Here, 
$\texttt{mc}<0$ indicates a loss barrier between the two models and hence poor connectivity;
$\texttt{mc}>0$ reveals lower loss regions between the models, which indicates poor training; and
$\texttt{mc}\approx0$ indicates well-connected models.

\subsection{Hessian Metrics Computation}
\label{app:hessian_calculation}
To analyze the second-order geometry of the loss landscape, we use Hessian-based metrics including top eigenvalues, trace, and spectrum. Given the high dimensionality of the parameter space, explicit computation of the Hessian matrix $H$ is computationally prohibitive. Therefore, we implement matrix-free methods based on Hessian-vector products (HVP), following the PyHessian framework~\citep{yang2021taxonomizing, yao2020pyhessian}.

\begin{itemize}
    \item \textbf{Top Eigenvalues:} The dominant eigenvalues ($\lambda_{max}$) are computed using the power iteration method. This iterative process identifies the directions of maximum curvature in the loss landscape, which are often associated with model sensitivity and sharp minima.
    \item \textbf{Hessian Trace:} The trace of the Hessian is estimated using Hutchinson's method. By computing the expectation of quadratic forms $v^T H v$ with Rademacher random vectors $v \in \{-1, 1\}^n$, we obtain an unbiased estimator of the average curvature.
    \item \textbf{Hessian Spectrum:} The Hessian Spectrum is estimated via the Stochastic Lanczos Quadrature to provide a global view of the curvature distribution. We follow the methodology in the NNCG~\citep{nncg}. To capture the heavy-tailed nature of the spectrum, the density is visualized using a symmetrical log-scale, allowing for the clear observation of eigenvalues spanning multiple orders of magnitude.
\end{itemize}

While standard Hessian analysis for architectures like ResNet, ViT~\citep{dosovitskiy2021imageworth16x16words}, or ConvNeXt~\citep{woo2023convnextv2codesigningscaling} typically relies on a single mini-batch for computational efficiency, our implementation for PINNs uses the entire set of residual collocation points. This ensures that the second-order information faithfully represents the full physics-informed loss surface rather than a stochastic approximation.

\subsection{Implementation Details of Boundaries}
\label{app:regime_boundary_implement}

The specific methodology for regime classification that we use is adapted from the taxonomic framework proposed in~\citet{yang2021taxonomizing}. We report boundary segmentation thresholds for different PDEs in Table~\ref{tab:boundary_thresholds}. No smoothing or post-processing is applied to the raw values shown in the heatmaps, ensuring that the regime plots faithfully reflect the original experimental measurements.

\begin{table}[ht!]
\centering
\caption{Detailed training and test-error thresholds used for regime-boundary extraction, grouped by model, PDE system, and optimization method.}
\label{tab:boundary_thresholds}
\resizebox{0.85\columnwidth}{!}{
\begin{tabular}{lllcc}
\toprule
\textbf{Model} 
& \textbf{Physical System} 
& \textbf{Training Strategy} 
& \textbf{Training Threshold} ($\boldsymbol{T_{\text{train}}}$)
& \textbf{Test Threshold} ($\boldsymbol{T_{\text{test}}}$) \\
\midrule

\multirow{16}{*}{PINN} 
& \multirow{5}{*}{1D Convection} 
                        & L-BFGS & $3.80 \times 10^{-3}$ & $0.434$ \\
&                       & ALM    & $3.34 \times 10^{-2}$ & $0.278$ \\ 
&                       & RoPINN & $2.97 \times 10^{-3}$ & $0.362$ \\ 
&                       & NNCG   & $3.55 \times 10^{-3}$ & $0.433$ \\ 
&                       & CL     & $2.29 \times 10^{-3}$ & $0.239$ \\ 
\cmidrule{2-5}

& \multirow{4}{*}{1D Reaction} 
                        & L-BFGS & $2.73 \times 10^{-2}$ & $0.469$ \\
&                       & ALM    & $7.30 \times 10^{-5}$ & $0.494$ \\ 
&                       & RoPINN & $1.84 \times 10^{-2}$ & $0.361$ \\ 
&                       & NNCG   & $2.46 \times 10^{-2}$ & $0.432$ \\ 
\cmidrule{2-5}

& \multirow{4}{*}{1D Wave} 
                        & L-BFGS & $9.91 \times 10^{-3}$ & $0.297$ \\
&                       & ALM    & $5.77 \times 10^{-3}$ & $0.300$ \\ 
&                       & RoPINN & $7.80 \times 10^{-3}$ & $0.306$  \\ 
\cmidrule{2-5}

& \multirow{3}{*}{1D Reaction--Diffusion} 
                        & L-BFGS & $4.46 \times 10^{-2}$ & $0.435$ \\
&                       & ALM    & $2.44 \times 10^{-4}$ & $0.384$ \\ 
&                       & RoPINN & $4.26 \times 10^{-2}$ & $0.411$  \\ 
\midrule 

\multirow{2}{*}{FNO} 
& 2D Poisson            & Adam  & $5.96 \times 10^{-2}$ & $0.192$ \\
\cmidrule{2-5}
& 2D Advection-Diffusion & Adam   & $6.67 \times 10^{-2}$ & $0.258$ \\
\midrule

\multirow{5}{*}{PINO} 
& \multirow{5}{*}{2D Darcy Flow} 
                        & Adam   & $0.211$               & $0.161$ \\
&                       & L-BFGS & $0.323$               & $0.163$ \\ 
&                       & ALM    & $0.125$               & $0.158$ \\ 
&                       & NNCG   & $0.125$               & $0.167$ \\ 
&                       & CL     & $0.107$               & $0.0973$ \\ 
\midrule

\multirow{1}{*}{NODE} 
& Nonlinear Damped Pendulum           & Adam   & $5.21 \times 10^{-5}$ & $1.98$ \\
\midrule

\multirow{5}{*}{PINODE} 
& \multirow{5}{*}{Nonlinear Damped Pendulum} 
                        & Adam   & $1.12 \times 10^{-6}$ & $1.74$ \\
&                       & L-BFGS & $3.20 \times 10^{-7}$ & $3.18$ \\ 
&                       & ALM    & $6.88 \times 10^{-9}$ & $1.31$ \\ 
&                       & NNCG   & $1.37 \times 10^{-7}$ & $2.19$ \\ 
&                       & CL     & $2.43 \times 10^{-7}$ & $3.47$ \\ 
\bottomrule
\end{tabular}
}
\end{table}

\subsection{Color Coding Protocol}
\label{app:color_coding}

\paragraph{2D Regime Plots.} Each cell in the 2D grid corresponds to a unique (physical parameter, data quantity) configuration. For each configuration, we train the model with multiple random seeds and report the seed-averaged metric value. The color of each cell is determined by mapping this averaged value to a truncated colormap (using the $[0, 0.8]$ portion of the full color range) through a robust percentile normalization. Specifically, for a metric value~$v$ at grid point~$(i,j)$, the color is assigned as
\begin{equation}
  c_{i,j} \;=\; \mathrm{colormap}\!\left(\frac{v_{i,j} - v_{\min}}{v_{\max} - v_{\min}}\right),
  \label{eq:color_norm}
\end{equation}
where $v_{\min}$ and $v_{\max}$ denote the 5th and 95th percentiles of the metric values across all grid points within that subplot, respectively. This percentile-based normalization ensures that extreme outliers do not dominate the color scale. Each subplot is independently normalized; different subplots may therefore have different color scales. Training loss and test error are visualized in separate panels, each with its own independent color scale. Values falling outside the $[v_{\min}, v_{\max}]$ range are clamped to the corresponding endpoint colors of the colormap.

\paragraph{Relative Improvement Heatmaps.}
Each cell shows the percentage change of the evaluated method relative to the baseline:
\begin{equation}
  \Delta_{i,j} \;=\; \frac{v^{\mathrm{base}}_{i,j} - v^{\mathrm{method}}_{i,j}}{v^{\mathrm{base}}_{i,j}} \times 100\%.
  \label{eq:relative_improvement}
\end{equation}
A diverging colormap is used, centered at zero, so that positive values (improvement over the baseline) and negative values (degradation) are visually distinguishable. All runs, including those exhibiting numerical instability or divergence, are included in the seed average without exclusion; hence, the displayed color at each cell faithfully reflects the raw averaged percentage change.

\paragraph{3D Loss Landscape Surfaces.}
To generate our plots, we use the filter-wise normalization presented in~\citet{li2018visualizing}. Given the parameter $\theta$ of a trained model, the resulting plots are computed by the following function: $f(\alpha,\beta)=\mathcal{L}(\theta+\alpha d_1+\beta d_2)$, where $d_1$ and $d_2$ are the two directions with maximum eigenvalues of $H_{\mathcal{L}}$. $\alpha$ and $\beta$ are uniformly sampled from $[-r,r]$, where $r$ controls the perturbation radius in the parameter space. Given that the value of $\mathcal{L}$ near the converged model is small, we use the log scale to make the trend more pronounced. 
The surface color encodes the log-scaled loss value
$\log \mathcal{L}(\theta+\alpha d_1+\beta d_2)$
at each point in the two-dimensional parameter slice, following the filter-wise normalization. A sequential colormap is used, with cooler colors corresponding to lower loss values
and warmer colors to higher loss values.

\paragraph{Hessian-based Visualizations.}
Two types of Hessian visualizations are used in this work.
\emph{Hessian phase plots} display the logarithm of the maximum Hessian eigenvalue,
$\log \lambda_{\max}(\mathbf{H})$, computed at the converged model for each configuration and averaged over random seeds. The color coding follows the same per-subplot robust percentile normalization as in~\Eqref{eq:color_norm}. \emph{Hessian spectral density plots} do not use a heatmap color coding; instead, the $x$-axis represents eigenvalue magnitude on a symmetric log scale, and the $y$-axis represents the estimated spectral density. Blue and orange curves distinguish the spectra before and after training, respectively. The spectral density is estimated via the Stochastic Lanczos Quadrature method, following the procedure described in Appendix~\ref{app:hessian_calculation}.

\section{Extended Analysis of Regimes Across Varying Physics}
\label{app:extened_regime_physics}

\begin{figure*}[ht!]
    \centering
    \begin{tikzpicture}
        \node[anchor=south west, inner sep=0] (img) at (0,0)
        {\includegraphics[width=1.0\textwidth]{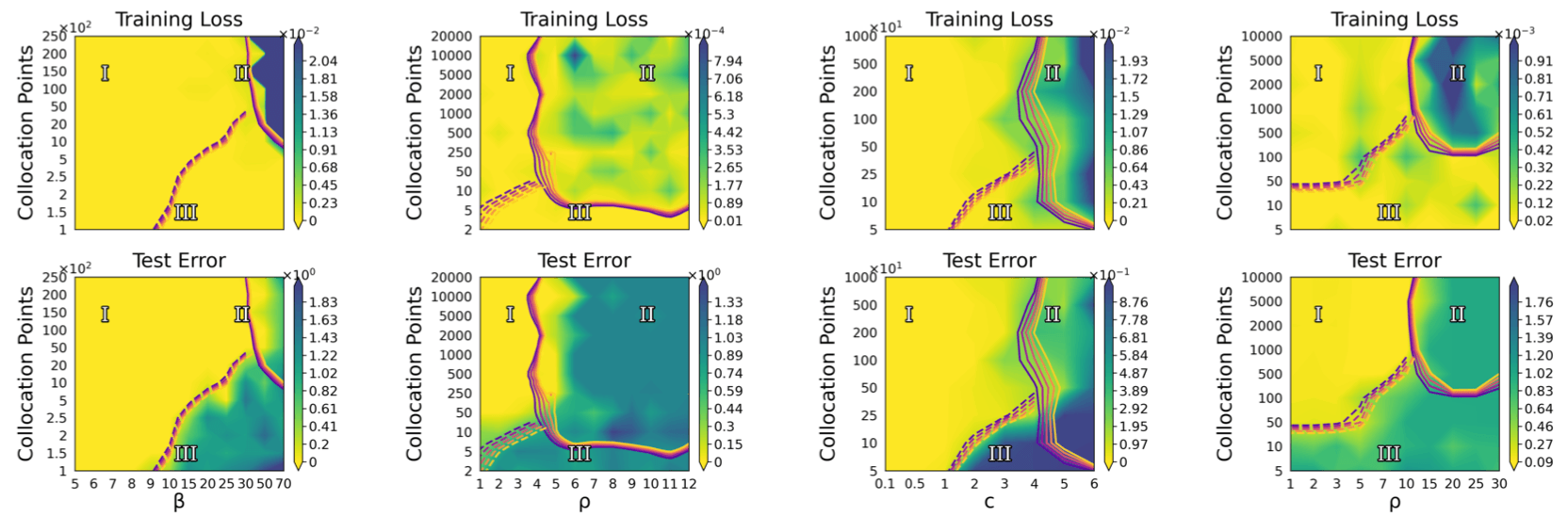}};

        \begin{scope}[x={(img.south east)}, y={(img.north west)}]
            \node at (0.11, -0.04) {\small (\emph{a}) 1D Convection};
            \draw[lightgray, dashed, line width=1pt] (0.235, -0.05) -- (0.235, 1.01);

            \node at (0.375, -0.04) {\small (\emph{b}) 1D Reaction};
            \draw[lightgray, dashed, line width=1pt] (0.5, -0.05) -- (0.5, 1.01);
            
            \node at (0.63, -0.04) {\small (\emph{c}) 1D Wave};
            \draw[lightgray, dashed, line width=1pt] (0.755, -0.05) -- (0.755, 1.01);

            \node at (0.885, -0.04) {\small (\emph{d}) 1D Reaction-Diffusion};

        \end{scope}
    \end{tikzpicture}

    \caption{
    Regime plots across varying physical systems using \textbf{a fixed PINN model trained with the ALM optimizer}. The evaluated PDE systems include (\emph{a}) the 1D convection equation with convection coefficient $\beta$, (\emph{b}) the 1D reaction equation with reaction coefficient $\rho$, (\emph{c}) the 1D wave equation with wave speed $c$, and (\emph{d}) the 1D reaction-diffusion equation with reaction coefficient $\rho$. Lighter (yellow) and darker (green) colors denote lower and higher training loss/test error, respectively.
    }
    \label{fig:regime_physics_alm}
\end{figure*}
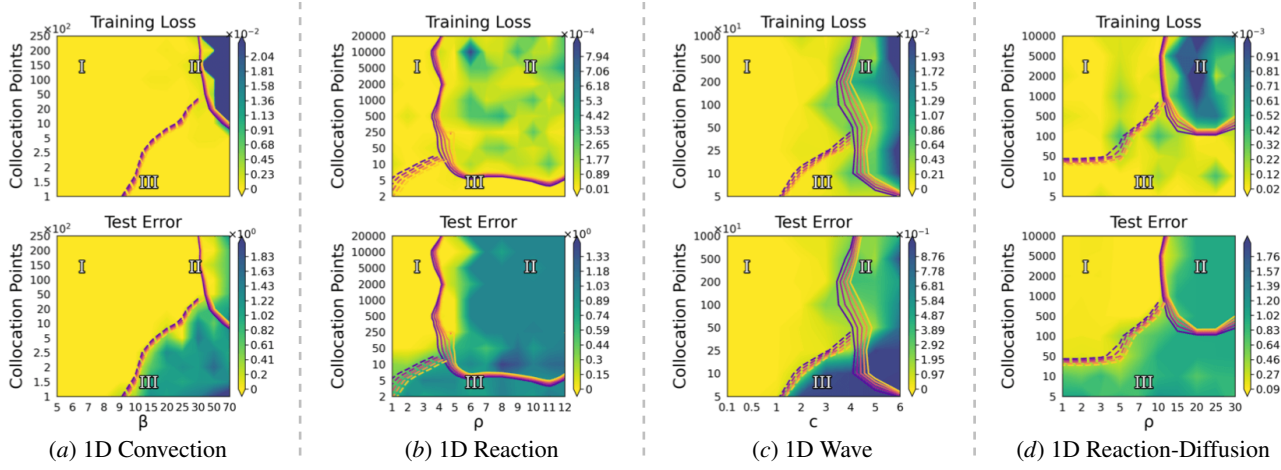

\begin{figure*}[ht!]
    \centering
    \begin{tikzpicture}
        \node[anchor=south west, inner sep=0] (img) at (0,0)
        {\includegraphics[width=1.0\textwidth]{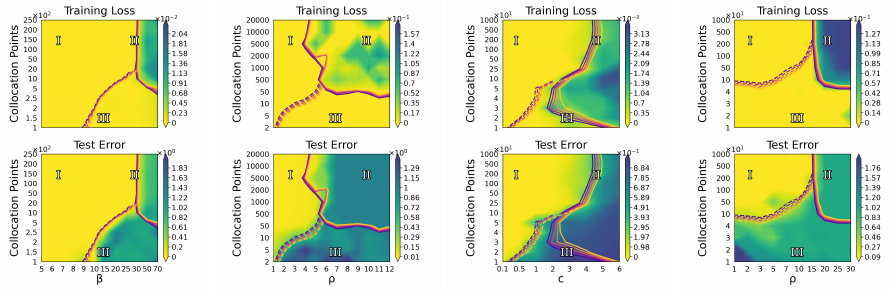}};

        \begin{scope}[x={(img.south east)}, y={(img.north west)}]
            \node at (0.11, -0.04) {\small (\emph{a}) 1D Convection};
            \draw[lightgray, dashed, line width=1pt] (0.235, -0.05) -- (0.235, 1.01);

            \node at (0.375, -0.04) {\small (\emph{b}) 1D Reaction};
            \draw[lightgray, dashed, line width=1pt] (0.5, -0.05) -- (0.5, 1.01);
            
            \node at (0.63, -0.04) {\small (\emph{c}) 1D Wave};
            \draw[lightgray, dashed, line width=1pt] (0.755, -0.05) -- (0.755, 1.01);

            \node at (0.885, -0.04) {\small (\emph{d}) 1D Reaction-Diffusion};
        \end{scope}
    \end{tikzpicture}

    \caption{
    Regime plots across varying physical systems using \textbf{a fixed PINN model trained with the RoPINN method}. The evaluated PDE systems include (\emph{a}) the 1D convection equation with convection coefficient $\beta$, (\emph{b}) the 1D reaction equation with reaction coefficient $\rho$, (\emph{c}) the 1D wave equation with wave speed $c$, and (\emph{d}) the 1D reaction-diffusion equation with reaction coefficient $\rho$.
    }
    \label{fig:regime_physics_ropinn}
\end{figure*}

\begin{figure*}[ht!]
    \centering
    \begin{tikzpicture}
        \node[anchor=south west, inner sep=0] (img) at (0,0)
        {\includegraphics[width=0.98\textwidth]{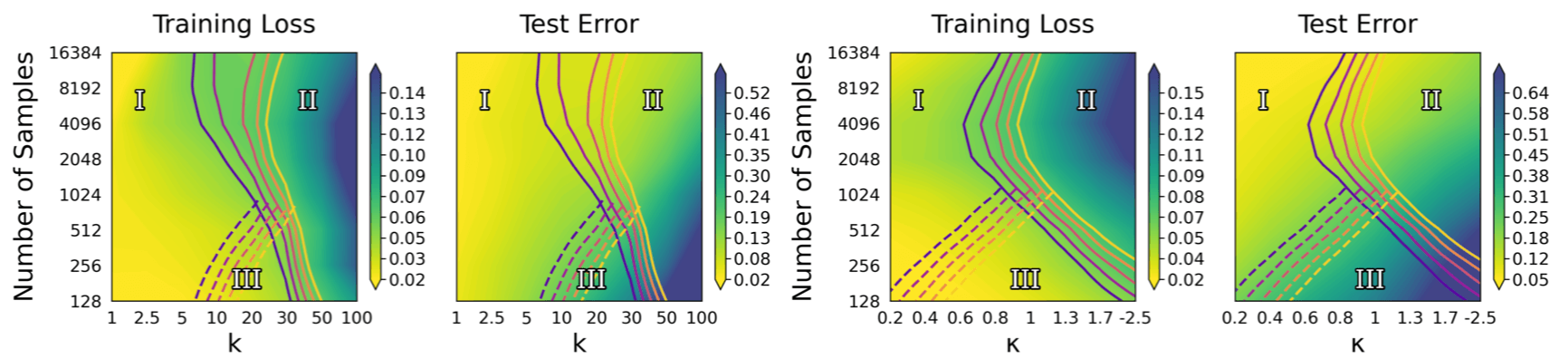}};

        \begin{scope}[x={(img.south east)}, y={(img.north west)}]
            \node at (0.25, -0.04) {\small (\emph{a}) 2D Poisson};
            
            \draw[lightgray, dashed, line width=1pt] (0.5, -0.05) -- (0.5, 1.01);

            \node at (0.75, -0.04) {\small (\emph{b}) 2D Advection-Diffusion};
          
        \end{scope}
    \end{tikzpicture}

    \caption{
    Regime plots across varying physical systems using \textbf{a fixed FNO model with the Adam optimizer}. The evaluated PDE systems include (\emph{a}) the 2D Poisson equation and (\emph{b}) the 2D advection-diffusion equation. 
    }
    \label{fig:regime_physics_fno}
\end{figure*}
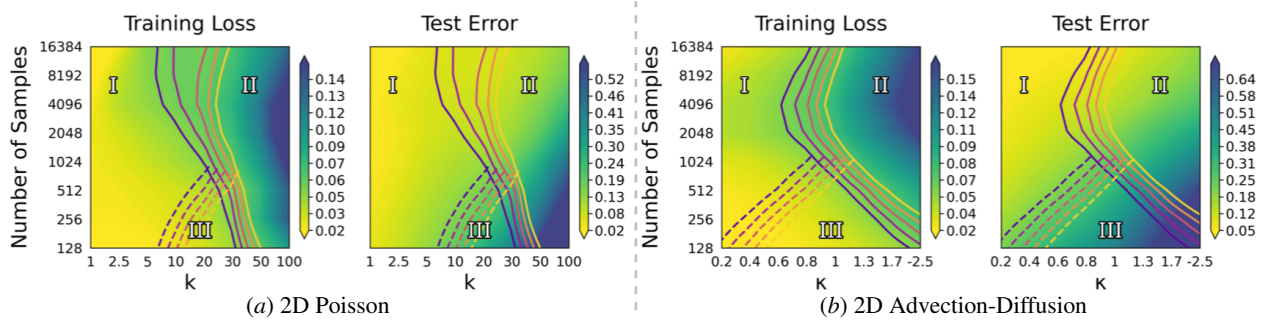

In this section, we provide additional regime-analysis results across varying physical systems. In addition to the main-text experiments, we evaluate PINNs with ALM and RoPINN on four representative PDE systems: the 1D convection, reaction, wave, and reaction-diffusion equations. Figures~\ref{fig:regime_physics_alm} and~\ref{fig:regime_physics_ropinn} further demonstrate that the coarse three-regime structure persists across different physical mechanisms, while the detailed geometry and location of the regime boundaries vary with the underlying PDE. As shown in Figure~\ref{fig:regime_physics_fno}, we also provide additional FNO experiments trained with Adam on the 2D Poisson and advection-diffusion equations. Compared with PINN-based models, the resulting regime maps exhibit smoother and less sharply separated transitions, but they still display consistent trainability and generalization patterns across varying physical difficulty and training-data regimes. Together, these extended results support the robustness of the regime-aware perspective across both physics-constrained and data-driven SciML paradigms.

\section{Extended Analysis of Possible Failure Mechanisms}
\label{sec:appendix_failure_modes}

In this section, we provide additional evidence to support our analysis of possible failure mechanisms in SciML optimization. Building on the loss-landscape and Hessian eigenspectra observations in the main text, we further investigate two possible failure mechanisms and three additional hypotheses commonly associated with optimization instability (basin jumping, loss barriers, and landscape aging). The following empirical results suggest that many classical explanations from conventional deep learning do \emph{not} fully account for the optimization difficulties observed in SciML, highlighting the distinctive geometric and physical characteristics of these models, as well as the need for improved model diagnostics.

\subsection{Deceptive Sharpness}
\label{app:extend_deceptive_sharpness}

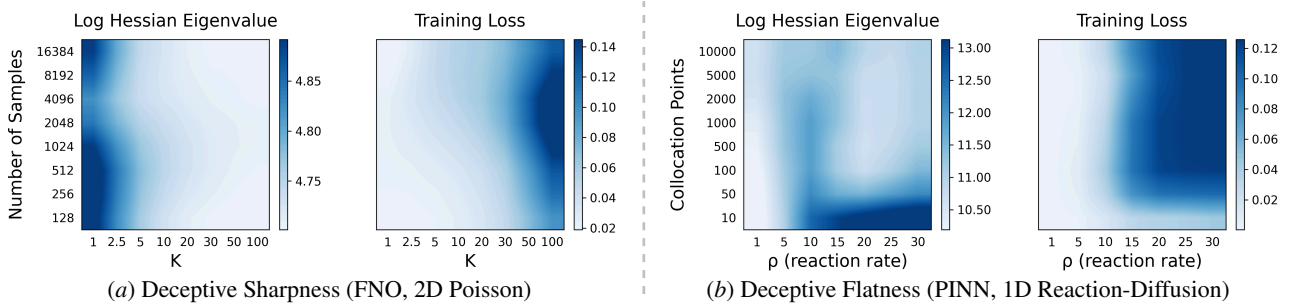
\begin{figure}[ht!]
    \centering
    \begin{tikzpicture}
        \node[anchor=south west, inner sep=0] (img) at (0,0)
        {\includegraphics[width=1\textwidth]{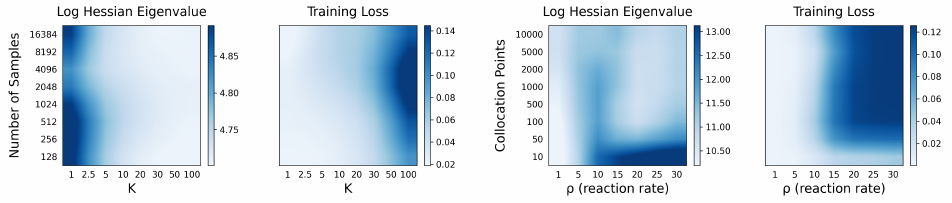}};

        \begin{scope}[x={(img.south east)}, y={(img.north west)}]
            \node at (0.25, -0.04) {\small (\emph{a}) Deceptive Sharpness (FNO, 2D Poisson)};
            
            \draw[lightgray, dashed, line width=1pt] (0.5, -0.05) -- (0.5, 1.01);

            \node at (0.75, -0.04) {\small (\emph{b}) Deceptive Flatness (PINN, 1D Reaction-Diffusion)};
          
        \end{scope}
    \end{tikzpicture}
    \caption{
    Examples of deceptive sharpness and deceptive flatness in SciML regime plots, illustrated through comparisons between log Hessian eigenvalues and training loss. (\emph{a}) FNO trained on the 2D Poisson equation, showing \textit{deceptive sharpness}, where regions with relatively low training loss still exhibit consistently large Hessian eigenvalues. (\emph{b}) PINN trained on the 1D reaction-diffusion equation, showing \textit{deceptive flatness}, where regions with smaller Hessian eigenvalues nevertheless correspond to poor optimization and large training loss. These results illustrate that Hessian sharpness alone does not reliably characterize trainability quality in SciML.
    }
    \label{fig:regime_hessian_train_loss}
\end{figure}

\begin{figure}[ht!]
    \centering
    \begin{subfigure}{0.42\textwidth}
        \centering
        \includegraphics[width=\linewidth]{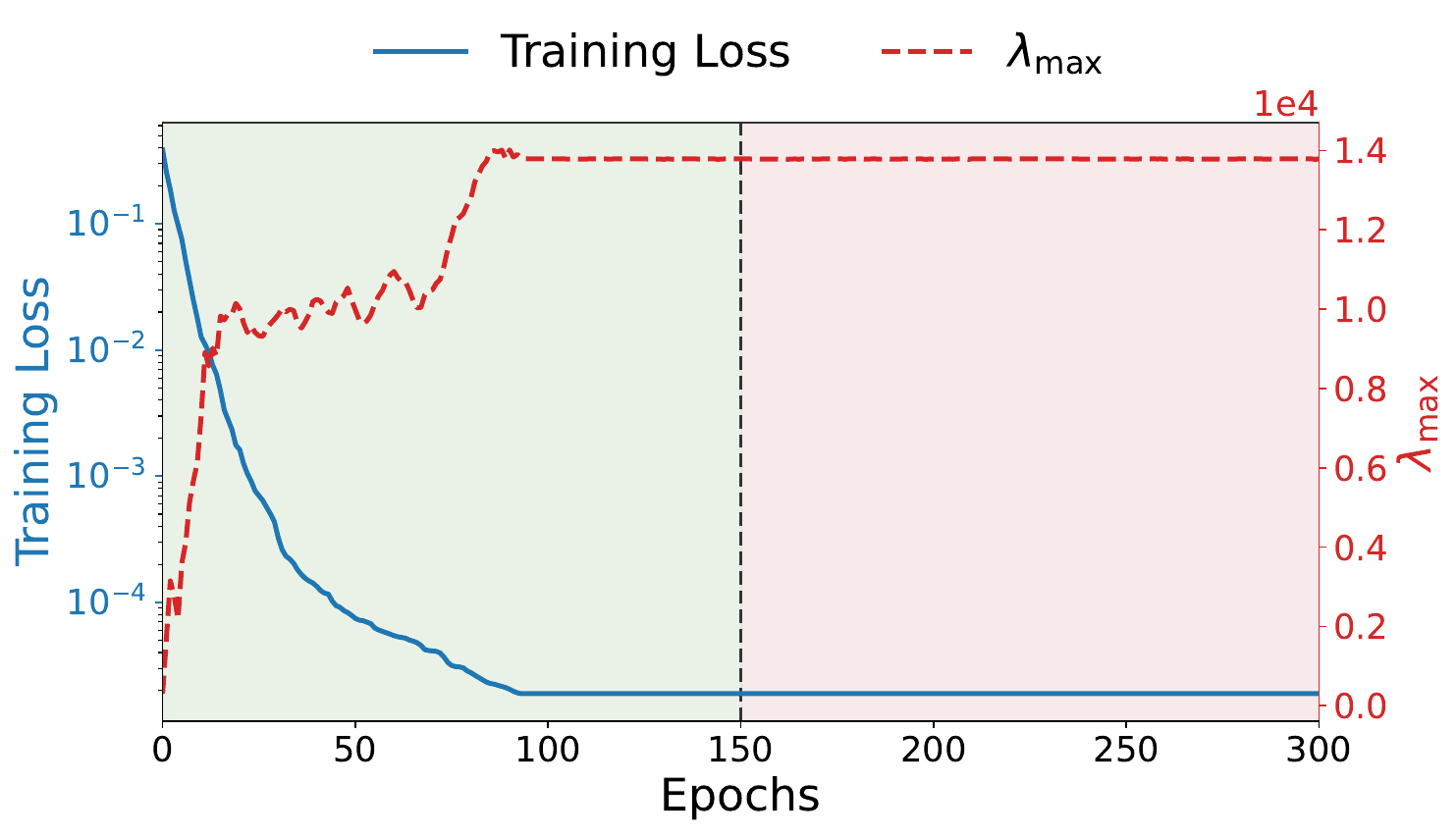}
        \caption{\small $N_f=100,\ \beta=5$}
    \end{subfigure}
    \hspace{3em}
    \begin{subfigure}{0.42\textwidth}
        \centering
        \includegraphics[width=\linewidth]{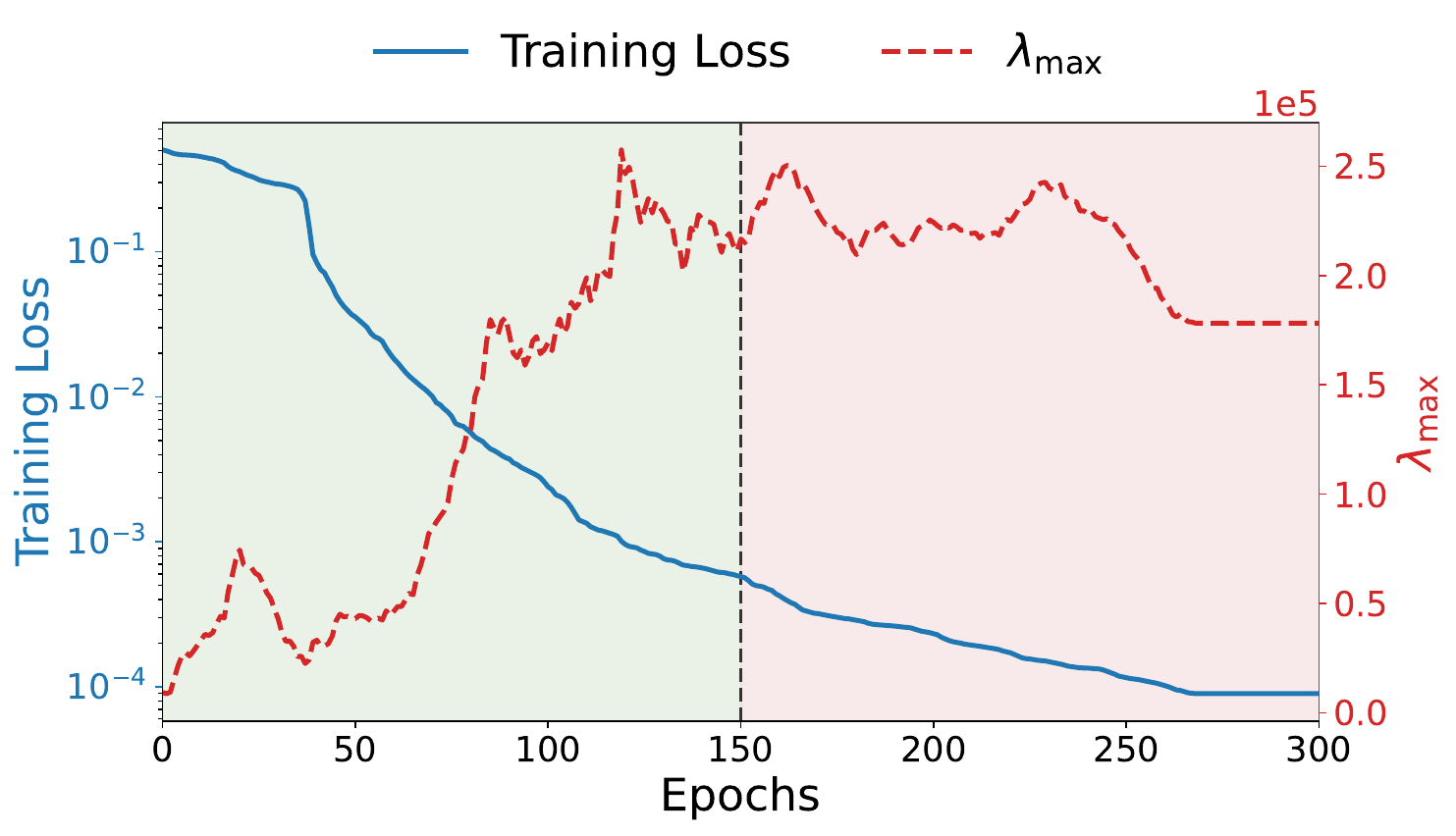}
        \caption{\small $N_f=100,\ \beta=50$}
    \end{subfigure}
    \vspace{1em}
    \begin{subfigure}{0.42\textwidth}
        \centering
        \includegraphics[width=\linewidth]{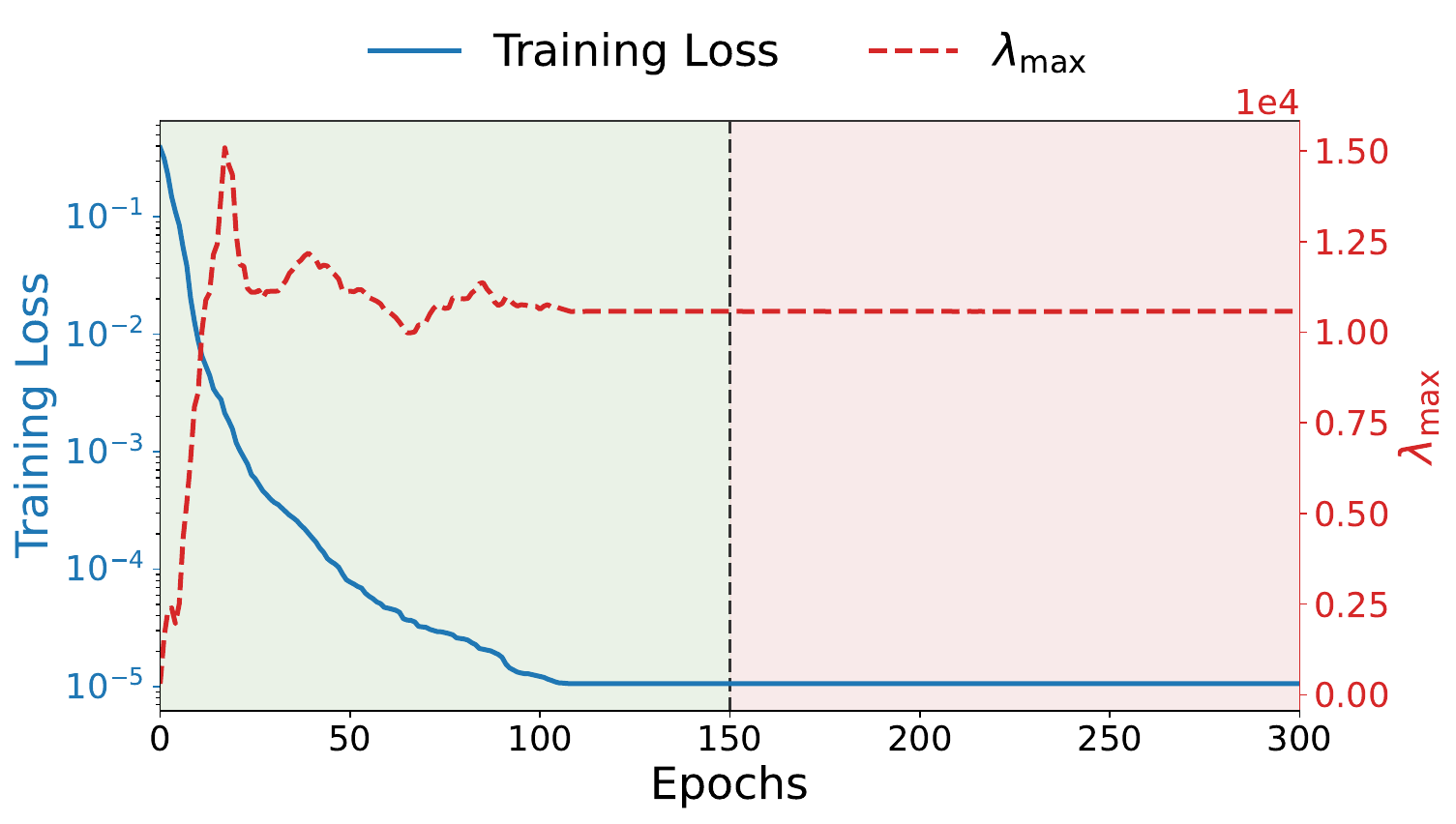}
        \caption{\small $N_f=1000,\ \beta=5$}
    \end{subfigure}
    \hspace{3em}
    \begin{subfigure}{0.42\textwidth}
        \centering
        \includegraphics[width=\linewidth]{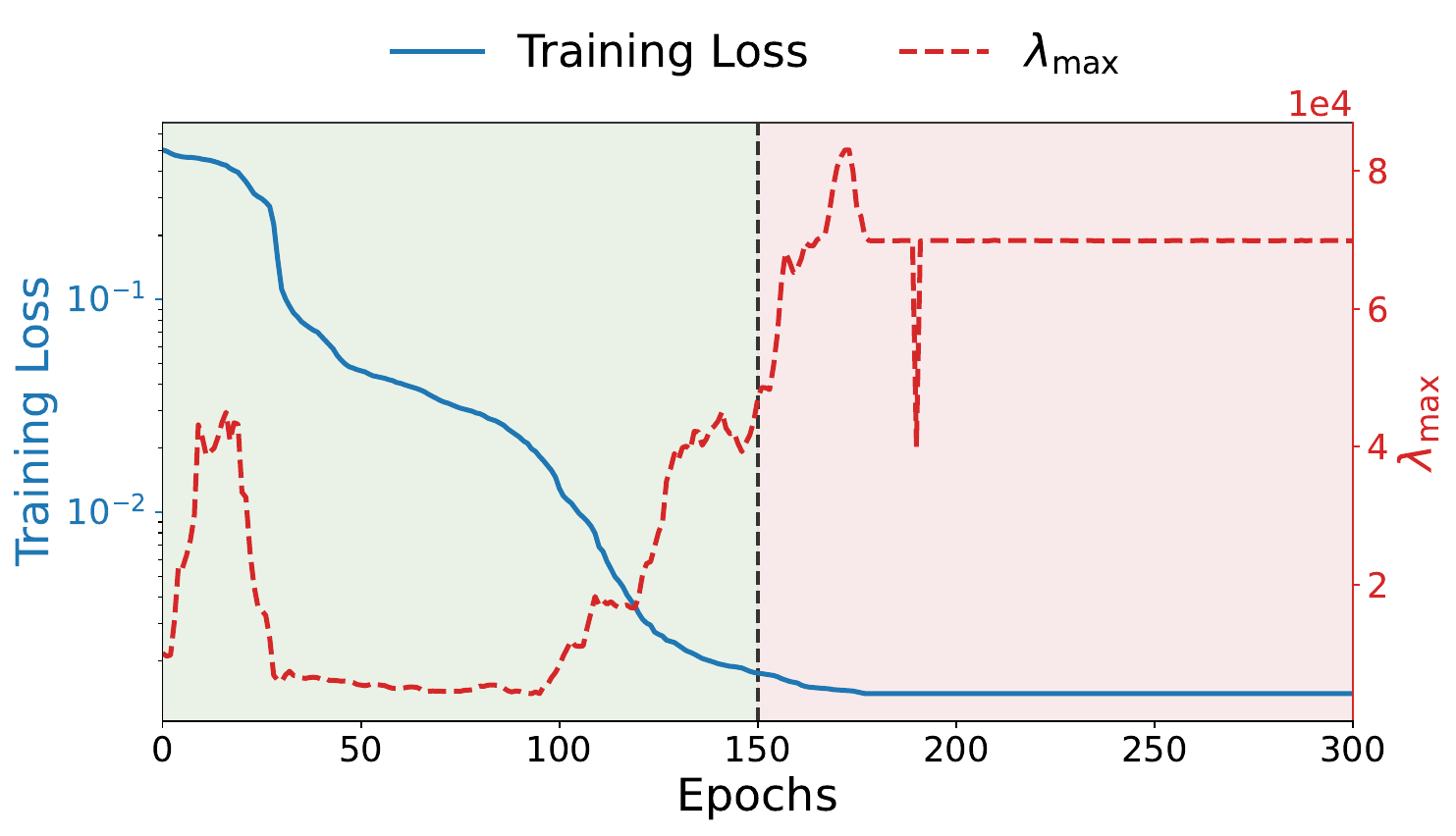}
        \caption{\small $N_f=1000,\ \beta=50$}
    \end{subfigure}
    \caption{
    Evolution of training loss and maximum Hessian eigenvalue $\lambda_{\max}$ during PINN training on the 1D convection equation under different physical and data regimes. The blue curve shows the training loss, while the red dashed curve shows the largest Hessian eigenvalue over training iterations. Panels vary the number of collocation points $N_f$ and the convection coefficient $\beta$. The shaded regions separate the early optimization phase and the later convergence phase, divided by the dashed vertical line. 
    }
    \label{fig:extend_increasing_sharpness}
\end{figure}

Here, we provide empirical evidence for deceptive sharpness. As shown in Figure~\ref{fig:regime_hessian_train_loss}(a), we observe an anomalous distribution of Hessian metrics when training FNO models on 2D Poisson datasets: some higher Hessian metric values appear in well-trained regions (Regime I), while lower values characterize poorly trained regions (Regime II and Regime III). Additional results on increasing sharpening are shown in Figure~\ref{fig:extend_increasing_sharpness} for PINNs trained on the 1D convection equation under varying collocation sizes and physical difficulty. Across all settings, PINN training exhibits progressive sharpening, where Hessian curvature increases substantially even as the training loss decreases. The sharpening effect becomes more pronounced in harder physical regimes with larger $\beta$, indicating increasingly ill-conditioned local geometry during optimization.

\subsection{Deceptive Flatness}
\label{app:extend_deceptive_flatness}

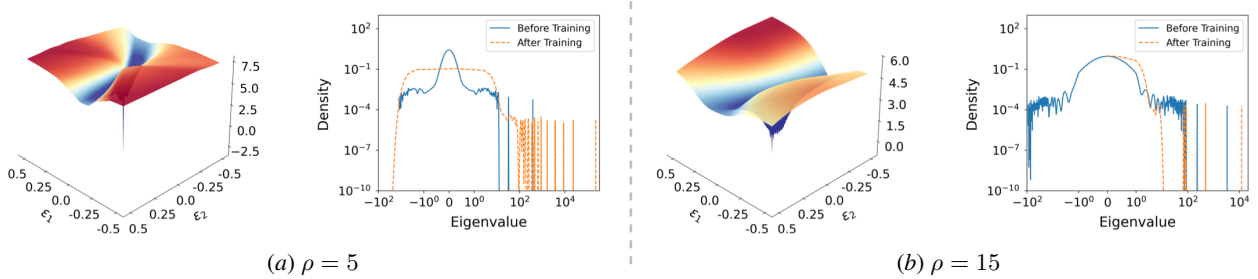
\begin{figure}[ht!]
    \centering
    \begin{tikzpicture}
        \node[anchor=south west, inner sep=0] (img) at (0,0)
        {\includegraphics[width=0.98\textwidth]{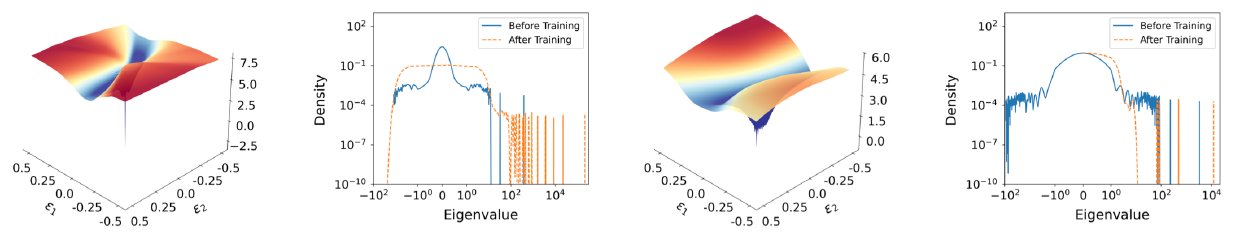}};

        \begin{scope}[x={(img.south east)}, y={(img.north west)}]
            \node at (0.25, -0.1) {\small (\emph{a}) $\rho=5$};
            
            \draw[lightgray, dashed, line width=1pt] (0.5, -0.1) -- (0.5, 1.01);
            
            \node at (0.75, -0.1) {\small (\emph{b}) $\rho=15$};
        \end{scope}
    \end{tikzpicture}

    \caption{
    3D loss landscapes and Hessian eigenspectra of PINNs trained on the 1D reaction-diffusion equation under different physical parameters. We use 1000 collocation points for training both models. Panel (\emph{a}) corresponds to $\rho=5$, while panel (\emph{b}) shows $\rho=15$. Although the $\rho=5$ setting exhibits a sharper minimum and larger Hessian eigenvalues after training, it achieves better optimization performance than the $\rho=15$ case. In contrast, the flatter landscape at $\rho=15$ is associated with broad high-loss plateaus and weaker descent structure, indicating that low curvature alone does not necessarily imply better trainability in SciML.
    }
    \label{fig:initial_flat_losslandscape}
\end{figure}

Here, we provide empirical evidence for deceptive flatness. We visualize the Hessian spectrum and training dynamics to characterize the pathological regions where optimization stalls.

\paragraph{Hessian Spectrum Analysis.}
Figure~\ref{fig:initial_flat_losslandscape} illustrates the evolution of the Hessian spectrum before and after training.
At initialization (blue curves), the spectrum exhibits a relatively symmetric structure centered around zero, indicating a diverse set of descent directions (negative eigenvalues) and ascent directions (positive eigenvalues).
However, as the optimizer interacts with the high-coefficient PDE ($\rho=20$), the spectrum undergoes a pronounced shift. The distribution becomes heavily biased toward positive values, while the magnitude of these eigenvalues remains small.

Quantitatively, we observe a significant reduction in the availability of descent directions. For the $\rho=20$ case, the proportion of negative eigenvalues drops to approximately 37\%. 
For context, this is substantially lower than standard deep learning benchmarks; for instance, a ResNet-18 model typically retains a negative eigenvalue proportion close to 50\%.
This spectral degeneracy suggests that the optimizer is not sitting in a saddle point with clear escape routes, but rather on a relatively flat, featureless plateau where gradient information is effectively vanished.

\paragraph{Regime Plots.}
To further investigate the relationship between flatness and solution quality, we analyze the training dynamics using 2D phase plots, as shown in Figure~\ref{fig:regime_hessian_train_loss}(b). These plots summarize the maximum Log Hessian Eigenvalue across different collocation point settings and PDE parameters ($\rho$).

The plots reveal distinct training regimes. Notably, the region corresponding to large PDE coefficients (e.g., $\rho=15, 20$) exhibits an anomalous behavior (highlighted in yellow/lighter colors in the heatmap). 
In this regime, the model exhibits \textit{low} Hessian eigenvalues—normally a signature of convergence to a flat minimum—yet sustains \textit{high} training loss.
This confirms our hypothesis of deceptive flatness: the L-BFGS optimizer is trapped in a region of vanishing curvature. Unlike the sharp minima found in successful runs (Region I) or the chaotic high-curvature regions, this ``silent'' failure mode provides no directional feedback, causing the optimizer to terminate prematurely in a non-optimal state.

\subsection{Loss Landscape Jumps between Basins} 
\label{app:unlikely_failure_model_1}

\begin{figure}
    \centering
    \includegraphics[width=0.5\linewidth]{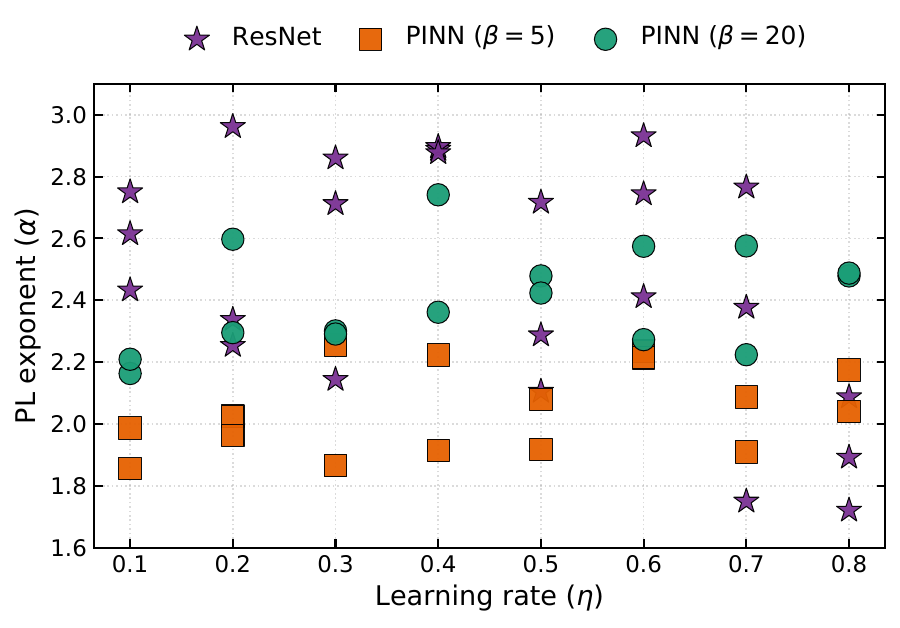}
    \caption{Comparison of lower-tail exponents $\alpha$ across learning rates $\eta$ for ResNet-18 and PINNs. Purple stars denote ResNet-18 trained with SGD, while orange squares and green circles denote PINNs for the 1D convection equation with $\beta=5$ and $\beta=20$, respectively, optimized using L-BFGS.}
    \label{fig:pl_exponent}
\end{figure}

We consider a potential hypothesis for the observed optimization difficulties: training fails because the optimizer unstably ``jumps'' between different basins of attraction, preventing convergence to a consistent minimum. This could be a plausible hypothesis, given the lack of basin connectivity observed in previous sections. The results are shown in Figure~\ref{fig:pl_exponent}. We follow the methodology of \citet{hodgkinson2022generalization} to quantify basin jumping and apply it to three settings: ResNet-18 (purple stars); a PINN for 1D convection with $\beta=5$ (orange squares); and a PINN for 1D convection with $\beta=20$ (green dots). The PINNs are optimized using L-BFGS, while ResNet-18 is optimized using SGD. Once the loss falls below a chosen threshold, we continue optimization for an additional number of iterations and record the step magnitudes $|W_{k+1}-W_k|$, following \citet{hodgkinson2022generalization}. We then compute the sequence $1/|W_{k+1}-W_k|$ for increasing $k$ and fit a power-law (PL) distribution using the \texttt{powerlaw} toolbox.

Figure~\ref{fig:pl_exponent} shows the fitted PL exponents $\alpha$, where smaller values of $\alpha$ indicate a higher tendency for basin jumping. We observe that both PINN settings generally exhibit larger $\alpha$ values than ResNet-18, suggesting fewer jumps. This is expected, as the better connectivity of the ResNet loss landscape may reduce the need for frequent basin transitions. However, it is noteworthy that the $\beta=20$ case (green dots) exhibits fewer jumps than the $\beta=5$ case (orange squares). This is somewhat surprising and suggests that basin jumping may be effectively suppressed by the severely degraded loss landscape at larger~$\beta$.

Moreover, we observe an abrupt drop in the PL exponent $\alpha$ for ResNet-18 as the learning rate increases to larger values, indicating more frequent transitions between basins. In contrast, the $\alpha$ values for the PINNs remain relatively stable across settings, further suggesting that PINN optimization exhibits fewer basin jumps than one might initially expect.

\subsection{Loss Landscape Barrier}
\label{app:unlikely_failure_model_2}

\begin{figure}[ht!]
    \centering
    \begin{subfigure}{0.3\textwidth}
        \centering
        \includegraphics[width=\linewidth]{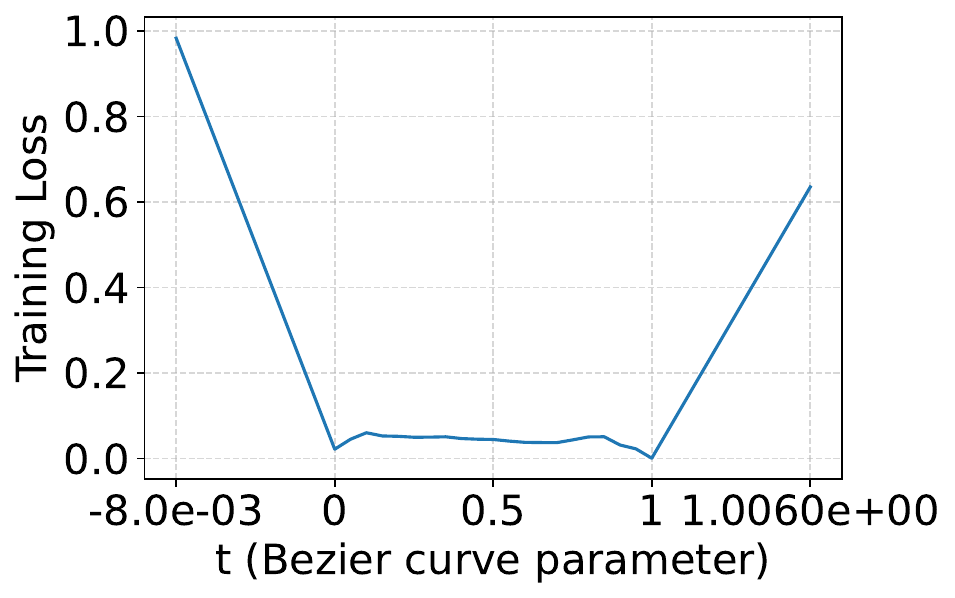}
        \caption{Training Loss}
    \end{subfigure}
    \hspace{4em}
    \begin{subfigure}{0.3\textwidth}
        \centering
        \includegraphics[width=\linewidth]{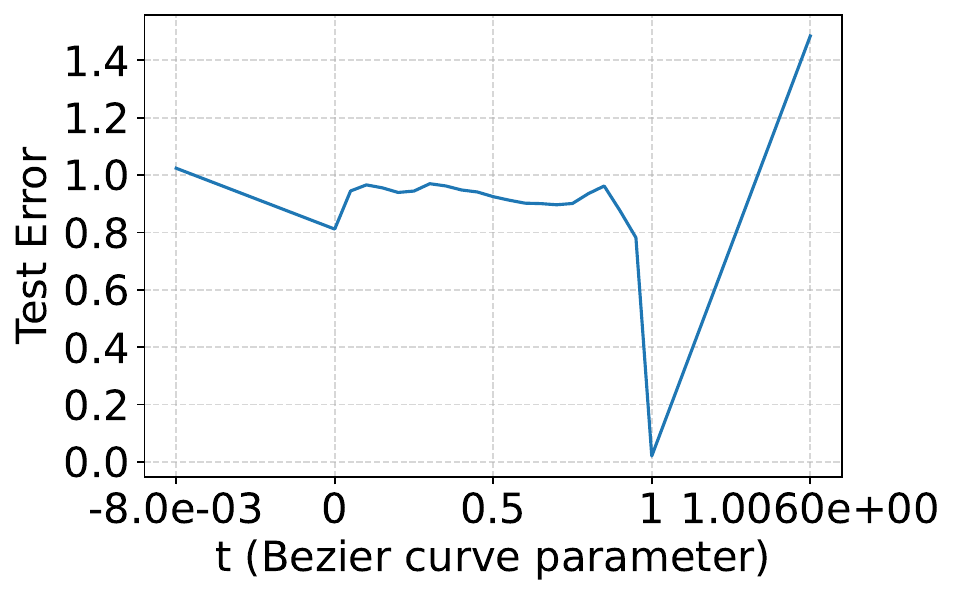}
        \caption{Test Error}
    \end{subfigure}
    \caption{Training loss and test error of PINN models on Bezier curve between L-BFGS ($t=0$) and ALM ($t=1$). We evaluate on 1D convection equation using $N_f = 10000$ and $\beta = 40$. As shown in (\emph{a}), the training loss landscape suggests that the L-BFGS and ALM solutions lie within the same basin. However, the corresponding test errors in (\emph{b}) differ significantly. This observation indicates that sharing a similar low-loss region does not necessarily imply comparable generalization performance; only sufficiently low training loss leads to physically accurate solutions.}
    \label{fig:curve_alm_pinn}
\end{figure}

Another possible loss landscape characteristic often investigated in CV is the loss landscape barriers; that is, the optimal solution and the current solution are separated by a high-loss barrier~\cite{draxler2018essentially,garipov2018loss}.
To further test whether the training difficulty of PINNs can be attributed to such a barrier within an otherwise correct loss basin, we conduct an additional evaluation that focuses on connecting a solution with coarse convergence to one with fine convergence. Specifically, we select the convection setting and archive model checkpoints at two distinct loss levels: a coarse-convergence level ($L \approx 10^{-2}$) in PINN with L-BFGS; and a fine convergence level ($L \approx 10^{-4}$) in PINN with ALM. The two checkpoints are at $t=0$ and $t=1$ in Figure~\ref{fig:curve_alm_pinn}. Although the coarse checkpoint already achieves a low training loss, it produces physically invalid solutions with large test error, failing to capture critical fine-scale structures. In contrast, further optimization to a much smaller loss is required to recover the correct physical behavior. This result indicates that the optimization difficulty does not arise from the need to climb a loss barrier. Instead, it reflects a more fundamental mismatch between loss minimization and solution fidelity in SciML, where low loss does not necessarily correspond to physically accurate solutions. 
In other words, optimization failures are not only due to optimizer inefficiency, but also arise from misleading loss landscapes that decouple training loss from solution accuracy.

\subsection{Loss Landscape Aging} 
\label{app:unlikely_failure_model_3}

\begin{figure*}[ht!]
    \centering
    \begin{tikzpicture}
        \node[anchor=south west, inner sep=0] (img) at (0,0)
        {\includegraphics[width=1.0\textwidth]{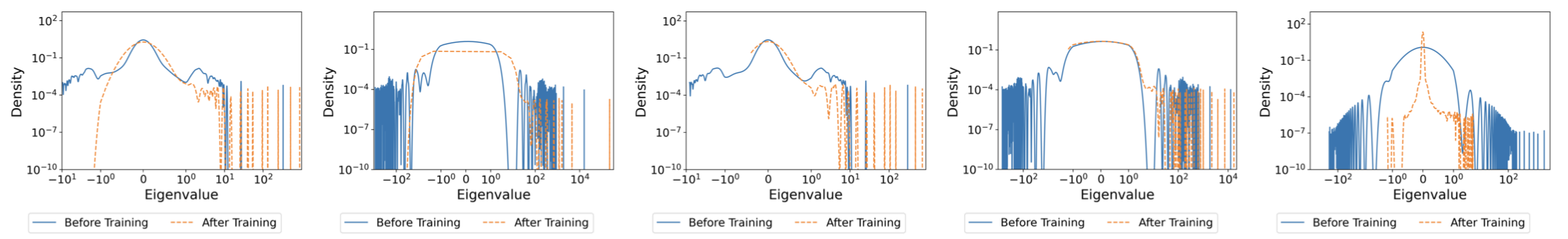}};

        \begin{scope}[x={(img.south east)}, y={(img.north west)}]

            \node[align=center] at (0.10, -0.2) 
            {\scriptsize (\emph{a}) PINN (Well-Trained)\\[-1mm]
            {\scriptsize $N_f=100,\ \beta=5$}};

            \node[align=center] at (0.30, -0.2) 
            {\scriptsize (\emph{b}) PINN (Over-Trained)\\[-1mm]
            {\scriptsize $N_f=100,\ \beta=50$}};

            \node[align=center] at (0.50, -0.2) 
            {\scriptsize (\emph{c}) PINN (Well-Trained)\\[-1mm]
            {\scriptsize $N_f=20\text{k},\ \beta=5$}};

            \node[align=center] at (0.70, -0.2) 
            {\scriptsize (\emph{d}) PINN (Under-Trained)\\[-1mm]
            {\scriptsize $N_f=20\text{k},\ \beta=50$}};

            \node[align=center] at (0.90, -0.2) 
            {\scriptsize (\emph{e}) ResNet-18};

            \draw[lightgray, dashed, line width=1pt] (0.20, -0.3) -- (0.20, 1.01);
            \draw[lightgray, dashed, line width=1pt] (0.40, -0.3) -- (0.40, 1.01);
            \draw[lightgray, dashed, line width=1pt] (0.60, -0.3) -- (0.60, 1.01);
            \draw[lightgray, dashed, line width=1pt] (0.796, -0.3) -- (0.796, 1.01);

        \end{scope}
    \end{tikzpicture}

    \caption{
    Hessian eigenspectrum density plots before and after training for PINNs under different training regimes and for ResNet-18. The PINN examples compare well-trained, over-trained, and under-trained regimes across different collocation budgets $N_f$ and PDE coefficient parameters $\beta$, while ResNet-18 serves as a representative CV model.
    }
    \label{fig:hessian_comparison_pinn_additional}
\end{figure*}
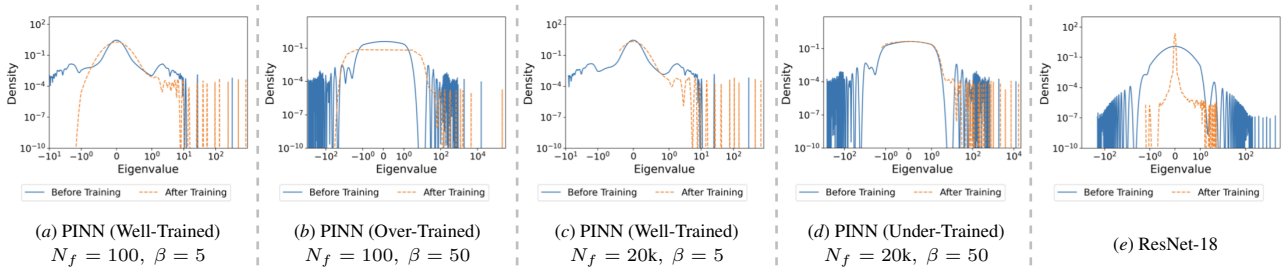

A third direction is ``loss landscape aging,'' a phenomenon where optimization dynamics slow down drastically during the final convergence phase~\citep{baity2018aging_reference}. This aging is characterized by the emergence of an increasing number of flat directions (i.e., Hessian eigenvalues approaching zero) after the optimizer has reached a wide, flat minimum. The optimizers' subsequent motion becomes diffusive and slow, hindering its ability to settle into the precise ``bottom'' of the basin. We contend, however, that this late-stage phenomenon does not explain the main optimization barrier observed in our SciML settings. Figures~\ref{fig:hessian_comparison_pinn_additional}(a-d) show that the PINN loss landscapes have a large number of non-zero eigenvalues and do not have a significantly higher density at zero. However, as shown in Figure~~\ref{fig:hessian_comparison_pinn_additional}(e), the density of eigenvalues of the Hessian matrix of CNN will decrease sharply when it extends from 0 outward in both directions. 
Since the ``aging'' hypothesis fundamentally presumes that the optimizer has already successfully navigated to a relatively low-loss, near-optimal region, we conclude that this hypothesis likely does not explain the hard training of the SciML settings we have considered. 
We note, of course, that this phenomenon also relates to increasing sharpening, as the bottom of the loss minimum remains sharper rather than flatter.

\end{document}